\documentclass[a4paper]{IEEEtran}          % twocolumn

\usepackage{balance}  % for  \balance command ON LAST PAGE  (only there!)

\usepackage[utf8]{inputenc}
\usepackage{alltt,xspace}
\usepackage{epsf,epsfig,float,graphicx}
\usepackage{color}
\usepackage[numbers]{natbib}
\usepackage{latexsym,amsmath,amsfonts,amssymb}

\usepackage{url}
\usepackage{verbatim}
\usepackage{moreverb}
\usepackage{theorem,enumerate}
\usepackage{pgfplots}
\usepackage{multirow}

\usepackage{listings} % pseudocodeg
\usepackage{color,colortbl}
\definecolor{dkgreen}{rgb}{0,0.6,0}
\definecolor{gray}{rgb}{0.5,0.5,0.5}
\definecolor{mauve}{rgb}{0.58,0,0.82}
\definecolor{lightGray}{gray}{0.9}

\newcommand{\removed}[1]{}

\newcommand{\ecomment}[1]{}
\newcommand{\hec}[1]{}

\newcommand{\lf}[1]{\texttt{#1}}
\newcommand{\cycle}{{Loop}}
\newcommand{\trans}{{Unnecessary Transshipment}}

\newenvironment{patternImpl}[1][DL-Axiom.]{\begin{trivlist}
  \item[\hskip \labelsep {\bfseries #1}]}{\end{trivlist}}

\newenvironment{dlquery}[1][DL-Query.]
{\begin{trivlist}
  \item[\hskip \labelsep {\bfseries #1}]}
  {\end{trivlist}}

\newcommand{\Ical}{{\cal I}}
\newcommand{\non}{\neg}

\newcommand{\finepatt}{\hfill~$\Box$}

\begin{document}
\title{	Semantic-based Anomalous Pattern Discovery in Moving Object Trajectories\thanks{This paper relies on the research presented in:
	P. Villa, E. Camossi, {\em A Description Logic Approach to Discover Suspicious Itineraries from Maritime Container Trajectories},
	In Proc. of GEOS 2011, LNCS 6631, p. 182-199. Springer-Verlag 2011.
	This research contributes to European Commission JRC action 41004 {\em Vessel and Container Surveillance}.}
}
\author{Elena Camossi
	\and
	Paola Villa
	\and
	Luca Mazzola %etc.
}

%\authorrunning{Short form of author list} % if too long for running head

\author{
\IEEEauthorblockN{Elena Camossi, Paola Villa, and Luca Mazzola}
\IEEEauthorblockA{
\\European Commission, Joint Research Center \textit{(JRC)}\\
Via Enrico Fermi 2749, Ispra, Varese, Italy -- I-21027\\
Email:
\{elena.camossi, paola.villa.pv, mazzola.luca\}@gmail.com\\
luca.mazzola@polimi.it
}
}

% \institute{Elena Camossi\at
%               European Commission, Joint Research Centre \\
% 	      via Enrico Fermi, Ispra, Varese (Va), Italy\\
%               %Tel.: +39-0332-78-5424\\
%               %Fax: +39-0332-78-9156\\
%               \email{elena.camossi@jrc.ec.europa.eu}           %  \\
% 	  \and
% 	    Paola Villa \at
%               European Commission, Joint Research Centre \\
% 	      via Enrico Fermi, Ispra, Varese (Va), Italy\\
%               %Tel.: +39-0332-78-9473\\
%               % Fax: +39-0332-78-9156\\
%               \email{paola.villa@jrc.ec.europa.eu}           %  \\
%               %\emph{via Enrico Fermi,Ispra, Varese (Va), Italy} of P. Villa  %  if needed
% 	  \and
% 	    Luca Mazzola \at
% 	      European Commission, Joint Research Centre \\
% 	      via Enrico Fermi, Ispra, Varese (Va), Italy\\
%               %Tel.: +39-0332-78-9473\\
%               % Fax: +39-0332-78-9156\\
%               \email{luca.mazzola@jrc.ec.europa.eu}
% }

\maketitle

%=====================================
\begin{abstract}

In this work, we investigate a novel semantic approach for pattern discovery in trajectories that, relying on ontologies, enhances object movement information with event semantics.
The approach can be applied to the detection of movement patterns and behaviors whenever the semantics of events occurring along the trajectory is, explicitly or implicitly, available.
In particular, we tested it against an exacting case scenario in maritime surveillance, i.e., the discovery of suspicious container transportations.

The methodology we have developed entails the formalization of the application domain through a domain ontology, extending the  Moving Object Ontology (MOO) described in this paper. Afterwards, movement patterns have to be formalized, either as Description Logic (DL) axioms or queries, enabling the retrieval of the trajectories that follow the patterns.

In our experimental evaluation, we have considered a real world dataset of 18 Million of container events describing the deed undertaken in a port to accomplish the shipping (e.g., loading on a vessel, export operation). Leveraging events, we have reconstructed almost 300 thousand container trajectories referring to 50 thousand containers travelling
along three years. We have formalized the anomalous itinerary patterns as DL axioms, testing different ontology APIs and DL reasoners to retrieve the suspicious transportations.

Our experiments demonstrate that the approach is feasible and efficient.
In particular, the joint use of Pellet and SPARQL-DL enables to detect the trajectories following a given pattern in a reasonable time with big size datasets.

%We have designed the Maritime Container Ontology, relying its formalization on container {\em events} that describe the deed undertaken to accomplish the shipping (e.g., loading on a vessel, export operation) in a port. Leveraging events, we reconstruct the container trajectories that are analysed to discover the suspicious cases.
%Anomalous itinerary patterns are formalized in the ontology as Description Logic (DLs) \cite{DBLP:conf/dlog/2003handbook} axioms, which can be used for querying the ontology and retrieving itineraries with anomalous behaviors.
%In this work we defined two example patterns:  the {\em cycle}, which occurs when a container comes back to its origin port before reaching its destination, and  {\em unnecessary transshipment}, which formalizes a container transfer that, from a mere route perspective, is useless to accomplish a shipment.

%Against these ones, we have run the DL queries implementing the anomalous patterns, testing different ontology APIs and DL reasoners.  % that can be run against the ontology.
%These data have been uploaded in the ontology and then queried running different reasoners and different query languages.

%To model the domain of containerised transportation,
%In the following, we have introduced the the Moving Object Ontology (MOO), a top-level ontology for modelling moving object trajectories.
% \ec{rewrite referring to the MOO}
\end{abstract}
%=====================================

%=====================================
\section{Introduction}\label{intro}
%=====================================

{\em Semantic trajectory} is a research trend that has recently emerged in Geographical Information Science and Spa\-tio-temporal Knowledge
Discovery~\cite{Alvares07,Guc08,spacca11,Spaccapietra,Yan2012}, to enhance the modelling and analysis of moving object data, e.g., GPS trajectories, mobile telephone streams, data collected from sensor networks.
In this domain, a moving object is an entity that changes position over time, such as a person that walks or cycles, a car, taxi or bus moving in a city, a vessel navigating by sea, etc.

In Semantic Trajectory, the goal is not the mere processing of the geographical trajectory for conventional GIS analysis, but the
{\em understanding} of the motion of the moving object with respect to the application of interest.
Therefore, the spatio-temporal modelling of object trajectory is enriched with semantic information that characterizes the
application context, such as the points of interest, like museums, schools, shops, etc., or the
annotation of parts of the trajectory to describe different movement behaviors, e.g., walking, cycling, driving.
Semantics enhances the analysis of data and facilitates the discovery of  semantically implicit patterns and behaviors \cite{Parent2013},
useful for abstracting the modelling domain and for inferring new knowledge.
In particular, the ontology-driven enrichment of moving object trajectories is a promising approach for the discovery of itinerary
\emph{patterns}~\cite{BaglioniMRTW09}, which can be applied for example to detect outliers in sequences of movements.

The analysis of moving object trajectories is a largely used tool in the field of maritime surveillance and security~\cite{camossi2012,etienne2012},
for fighting commercial frauds \cite{ctrf} and for enforcing the supply chain security to fight smuggling, counterfeiting and drug traffic.
Beyond its importance from an economic and citizen security perspective, supply chain monitoring is a challenging application scenario, in particular because  the number of containerized shipments to verify is enormous. Indeed, containers are used to ship the 25\% of world trade cargo, and even if recent legislation imposes to increase the inspections rate, currently less than 2\% of containers can be physically checked without causing expensive delays in the good trade chain.
Furthermore, 90\% of containers, i.e., 19 millions per year, travel by sea, with an estimated growth to reach 27 million by 2020. This, combined with the complexity of the shipping operations and with the number of
subjects involved, makes containerized transport particularly suitable to conceal illegal or hazardous materials.

In such a complex domain, effective Risk Analysis tools are essential to help Customs authorities identify effective suspicious transportations.
Route-based risk indicators (RRI), for example, target high risk consignments of goods by evaluating the trajectories of cargos, ships and containers.
RRIs analyse spatial information such as the ports where a container has been loaded and
discharged, the logistic of transshipment operations, and the actual route followed by a container.
RRIs support more traditional risk factors, such as the name of the consignee, the carrier, the value of transported goods.

In this work, we describe a novel methodology for semantic pattern discovery that relies on ontology and describe the tests we have run in the maritime surveillance scenario to detect suspicious containerized transportations. The approach we propose relies on a top-level ontology for modelling moving object trajectories, namely the Moving Object Ontology (MOO),
that has to be extended to represent the properties of the specific application domain.
On top of this formalization, movement patterns of interest may be defined as Description Logic (DL)\cite{DBLP:conf/dlog/2003handbook} axioms.
The ontology instances that satisfy the axioms represent the trajectories with the modelled movement behaviour.

In our test scenario, we have defined a knowledge base for the domain of maritime containers, namely the Maritime Container Ontology (MCO)~ \cite{DBLP:conf/geos/VillaC11}, and modelled {\em anomalous} container patterns that describe suspicious movement behaviors. We have run a set of experiments translating the axioms into DLs queries, that can be easily tested with different ontology APIs and reasoners on the populated ontology, retrieving the suspicious shipments that follow the defined patterns.
%The MCO formalization is rooted on container and vessel {\em events}, which describe the deeds undertaken on containers to accomplish shipment operations and the arrival and departure operations of vessels in ports.

For our tests, we consider two suspicious pattern examples, the proposed formalization can be extended to any number of patterns. The patterns we considered are {\em \cycle} and {\em \trans}, and  are well known in maritime risk analysis. They formalize irregular behaviors involving not only containers but also different vessels, because usually more than one vessel is used to accomplish a container shipment and containers are moved from one vessel to another during transshipment operations.
Such patterns are complex enough to show the potentialities of the semantic approach we propose, and are a step forward with respect to existing approaches proposed in the literature to detect patterns in moving object trajectories \cite{Baglioni08}.
However, despite they apparent complexity, they may be successfully discovered by integrating the knowledge of the locations where the events occur and the event semantics.

The methodology we propose can be applied in every context where the event semantics can be explicitly described with respect to {\em STOPs} or {\em MOVEs}~\cite{Spaccapietra}: specifically, STOPs are the places where a moving object stays for a minimum amount of time, while MOVEs are the subtrajectories between consecutive STOPs.
In our application scenario, we modelled STOPs and enriched them semantically with information on container and vessel {\em events}.
These ones describe the deeds undertaken on containers to accomplish shipment operations and arrival and departure operations of vessels in ports.

The advantages of the semantic approach we propose in this paper are twofold. First, abstracting the properties of the domain to high-level semantic concepts,
it simplifies the reasoning. For instance, every carrier company represents information on events using its own vocabulary but, within the ontology,
we can abstract from different vocabularies and reason on generic categories of events that are relevant for the application, such as transshipment events.
Moreover, our formalisation relies on DLs, a family of formal knowledge representation
languages used to describe and classify concepts and their instances, that combine
good expressivity and good computational properties, supporting the practical feasibility of the approach.
Indeed, knowledge representation systems based on description logics have been proven useful for structurally representing the terminological knowledge of an
application domain. Compared with first-order logic, DLs achieve a better trade-off between the
computational complexity of reasoning and the expressiveness of the language. DLs are briefly introduced in Section~\ref{sec:dls}.

The research presented in this paper relies on a previous work~\cite{DBLP:conf/geos/VillaC11}, where we introduced the MCO design and the application of axioms for anomalous patterns discovery in container itineraries. With respect to~\cite{DBLP:conf/geos/VillaC11}, in this work: (1) we
abstract from the application domain to define a
methodology for semantic pattern discovery that can be applied in other domains involving moving object trajectories; (2) we define DL-queries, semantically equivalent to ontology axioms, for the efficient retrieval of trajectories that verify the axioms conditions; (3) we run an extensive
experimental evaluation on a real world dataset to test the feasibility of the approach.

In our experiments, we have tested different DL reasoners, i.e., Hermit~\cite{hermit}, Pellet~\cite{pellet}, and FaCT++~\cite{fact},
and two of the most used API for DL querying: OWL-API~\cite{Horridge2011-owlapi} and SPARQL-DL API~\cite{Sirin07sparqldl}, and run the queries implementing the anomalous patterns against four ontologies of increasing size.
These have been populated with data taken from a dataset of eighteen million container events, preprocessed to define three hundred thousand container shipments. We have verified that the implementation solution combining SPARQL-DL API and Pellet achieves the maximum query language expressivity with the best performance, enabling to test a suspicious patterns in few minutes.

In the following, we use the term {\em container trajectory} to refer the spatial trajectory a container follows along a shipment, while with the term {\em itinerary} we refer to the same trajectories, semantically annotated with information on the events that occur during the shipment.

The rest of the paper is organized as follows.
We first provide the background of this research, discussing recent work on Semantic Trajectories
in Section~\ref{sec:related} and introducing the basic concepts of DLs in Section~\ref{sec:dls}.
In Section~\ref{sec:methodology}, we present the methodology we propose for the discovery of patterns and behaviors
in moving object trajectories, that we apply to the domain of containerized transportation in the next sections:
we describe the domain knowledge base for maritime container MCO (Section~\ref{sec:mco}) and give the description logic formalisation of suspicious
container itineraries (Section~\ref{sec:axioms}). Before introducing the experiments we have run in this domain (Section~\ref{sec:exp}), in Section~\ref{sec:semtools}
we compare the different tools and API for ontology querying that we have evaluated for our experimental evaluation.
Finally, in Section~\ref{sec:conclusion} we  discuss the potential development and the shortcomings of the approach we are proposing, concluding the paper.

%=====================================
\section{Semantic Trajectories}
\label{sec:related}
%=====================================
Most of the research on Semantic Trajectory has originated by the community grown within the FP6 project
GeoPKDD~\cite{geopkdd}, %~\footnote{Geographic Privacy-aware Knowledge Discovery and Delivery},
whose original focus was on privacy aware exploitation of spatio-temporal data. To continue the investigation on the discovery of knowledge and exploitation of moving object data, GeoPKDD has been followed first by MODAP~\cite{modap}
%~\footnote{Mobility, Data Mining, and Privacy} and
and more recently by SEEK~\cite{seek}. %~\footnote{SEmantic Enrichment of trajectory Knowledge discovery}.
The same community has recently presented a  survey of the research on this area~\cite{Parent2013}.
Among the active initiatives aiming at boosting the research on moving object modelling, analysis and visualization, a notable contribution has originated also by the COST Action MOVE~\cite{move}.

Another recent overview has been presented by Spaccapietra and collaborators~\cite{spacca11}, the same group that  originally proposed the first conceptual model for the representation of semantics in trajectories~\cite{Spaccapietra}, which has become a reference model for trajectory data analysis (for example,~\cite{Alvares07,Guc08,BaglioniMRTW09,bogornyIJGIS2009} refer to this model).
This model relies on the conceptualization of STOPs and MOVEs in trajectories: a STOP is an interesting
place in which a moving entity has stopped or reduced significantly its speed for a sufficient amount of time, likely to accomplish some activity; a
MOVE is any subset of the object trajectory between consecutive STOPs, and can be classified, for example, with respect to the type of moving (e.g., running, cycling, driving) or by the mean of transportation used to move.

Most of the research advances on trajectories and semantics may be broadly classified among three research areas:
Spatio-temporal Data Modelling for the representation of semantic trajectories;
Knowledge Discovery from Data (KDD) for semantic trajectory mining; and Geographic Visualization and Visual Analytics for semantic trajectory visualization.
In the rest of the section, we first overview work on semantic trajectories falling in the research areas above; then, we conclude discussing how our approach differs from the existing state of the art.

%----------------------------------------------
\subsection{Representing Semantic Trajectories}
%----------------------------------------------

For the representation and modelling of semantic trajectories, we can distinguish two different approaches: a traditional one that includes
moving object semantics since the phase of data design, and a-posteriori approach in which trajectories are annotated by analyzing its raw features,
such as the speed of the moving object or the intersection of the object trajectory with Places Of Interest (POI) previously extracted from the corresponding geographical layer.

The first approach is adopted in~\cite{Christophe}, where the authors introduce an algebraic model that represents a
spatio-temporal trajectory as an Abstract Data Type (ADT) that
encapsulates the semantic dimension. A series of trajectory states is
potentially observed and measured, and the ADT
representation combines a formal definition with manipulation operations, allowing the user to formulate queries on the semantics of the
spatio-temporal trajectory data type. Close to this approach we can account also the work of Pfoser et al.~\cite{Pfoser2003}, that generate synthetic datasets of semantic trajectories. %\ec{add something on the approach they use}

The second approach, which can be also referred to as (semantic) {\em segmentation} of trajectories,  or {\em episodes} identification, is more frequent in the literature.
The resulting representation is compliant to the model defined by Spaccapietra et al.~\cite{Spaccapietra} whenever interesting places, activities or means of transportation are identified to annotate the STOPs and MOVEs of the trajectory.
In particular, STOPs, somewhere called stay points, semantic places or locations, %\ec{add references},
distinguish the different {\em episodes}, i.e., the significant segments of a trajectory  that identify different phases of the object movement and can be assigned a clear semantics, relevant for the application domain.

Information on candidate STOPs is often encoded in the underlying geographical representation. For example, Cao et al.~\cite{Cao2010} and Guc et al.~\cite{Guc08} select STOPs from
pre-encoded POIs crossing the moving object trajectory. Alvares et al.~\cite{Alvares07} apply a similar approach, but selecting the Regions of Interest (ROI) in which the moving object stays for more than a given time, a temporal threshold that can differ for each ROI and is encoded within the ROI representation at a semantic level.
Cao et al.~\cite{Cao2010} give also a ranking of the top-k significant locations for each trajectory. The significance of locations for a user is discussed also by Zheng et al.~\cite{Zheng2009}, who adopt a hierarchical approach to detect important places and typical travel sequences from user trajectories.

Other works infer STOPs evaluating only the raw features of the trajectory, for example, the time the moving object does not move along the trajectory and the distance
between these stops~\cite{Zheng2011}, the change of speed ~\cite{Palma2008} or direction~\cite{Rocha2010}.

The two approaches can be combined, validating and correcting the geographical position of the STOPs resulting by the trajectory features processing with contextual information, like in the work by Yan et al.~\cite{Yan2012,Yan2011}.
Moreover, Yan et al.~\cite{Yan2012,Yan2011} abstract from  the requirement of a specific application domain using POI, ROI and Lines of Interest to annotate STOPs, and enabling to annotate also MOVEs, both as activities, such as walking, driving, cycling, and transportation modes, like bus, car, taxi, etc.

Annotation of MOVEs is also addressed by Yan et al.~\cite{SeTraStream}, who realize {\em online} identification of episodes by detecting the alteration of patterns within the trajectory. The trajectory segmentation adopts an existing approach for the discovery of trends that evaluates correlation coefficients, and incorporates also modules for trajectory cleaning and compression.
The episode tagging is done at a second stage by a classification model trained on trajectory features collected during the online segmentation, such as distance, duration, density, speed, acceleration, heading.

Annotation of MOVEs is also manually assisted by the visual tool developed by Guc et al.~\cite{Guc08}.
The work of Wannous et al.~\cite{Wannous2013} is a case of MOVEs annotation for animals trajectories, specifically seals', to distinguish travelling states (e.g., travelling, resting, foraging). They adopt ontologies to integrate the time knowledge to infer the different travelling states, which differentiate on duration and are defined in term of temporal axioms. Zhu et al.~\cite{Zhu2012} segment GPS trajectories of taxis to infer the taxi status, i.e., free, occupied or parked.
Wang et al. in \cite{Wang06} apply clustering on whole trajectories to distinguish among different trajectory types (e.g., pedestrian, vehicles) and activities (e.g., walking, cycling). In this case the labelling is done on an entire trajectory. The result of the clustering is used in particular to infer the structure of the scene in which the objects are moving.

Clustering is also used by Cao et al.~\cite{Cao2010} for the extraction of semantic locations and by Palma et al.~\cite{Palma2008}, who adopt spatio-temporal clustering to classify trajectory with respect to their speed.

Finally, van Hage et al.~\cite{vanHage2009} present an interesting approach for modelling and analysing ship trajectories for early time awareness for
Maritime Surveillance and Security, which takes into account the semantics of the trajectories.
Taking in input Marine Automatic Identification System (AIS) messages sent by ships, they build trajectories and segment them by detecting the significant events that represent changes in ship behaviour, such as speeding up, anchored, stopped.
Reasoning rules for event labelling are specified in SWI-Prolog, and the geographical knowledge relies on the GeoNames\footnote{www.geonames.org} ontology.

\subsection{Knowledge Discovery and Exploitation of Semantic Trajectories}
%-------------------------------------------------------------------------

As we have seen, some of the methods described above \cite{Wang06,Cao2010,Palma2008} adopt data mining, clustering in particular, for the semantic annotation of trajectories.
However, there are also approaches that exploit semantic trajectory for knowledge discovery, in particular movement patterns. In this area, several works have been published by the communities collaborating within the project GeoPKDD and its followers.

Alvares et al.~\cite{Alvares2007b} and Moreno et al.~\cite{MorenoTRB10} take semantic trajectory with annotated STOPs and MOVEs and extract moving patterns considering also background geographical information.
Bogorny et al. in~\cite{Bogorny2011} present Weka-STPM, a data mining toolkit for geographical data that takes trajectories with annotated POIs and performs episode recognitions as pre-processing for analysis and visualization.
Bogorny et al. in~\cite{bogornyIJGIS2009,Bogorny2010} formalize the idea of semantic trajectory pattern mining to boost data preprocessing and to mine data at a higher abstraction level. They discuss in particular the discovery
of frequent and sequential patterns and  association rules from trajectories.
Relying on the results presented in~\cite{Alvares07,Palma2008}, they preprocess trajectories to annotate STOPs and MOVEs. Then, mining can be applied directly on the annotated dataset.

Ying et al.~\cite{Ying2010} compute similarity of user trajectories, taking into account trajectory semantics. The same authors in ~\cite{Ying2011} rely on  user behaviour in similar clusters to predict the next location in a semantic trajectory.

Baglioni et al.~\cite{Baglioni08,BaglioniMRTW09} represent annotated trajectories in an ontology encompassing also geographical and application domain knowledge.
Different kinds of STOPs are considered, and temporal knowledge is used to discriminate among them.
Afterwards, they use ontology axioms to infer behaviour al patterns.

Similarly to~\cite{Baglioni08,BaglioniMRTW09}, Yan et al.~\cite{YanQuery08} use an ontological approach for the representation of semantic trajectory.
They define three different ontology modules for representing geometry, geography and the requirements of the application domain and apply their
approach to the application case of traffic management.
The geometric modules includes a Trajectory Ontology compliant with the model defined by the same authors
in~\cite{Spaccapietra}. In their approach, the ABox of the ontology, containing the ontology instances, is stored in a database, specifically
Oracle extended with Oracle Semantics, which includes the OWLPrime language, a DL subset, for ontology representation, querying and inference.

Based on space time ontology and events approach, Boulmakoul et al.~\cite{DBLP:journals/corr/abs-1205-1796} propose a generic meta-model for
trajectories of moving objects to allow independent applications processing trajectories data benefit from a high level of interoperability,
information sharing as well as an efficient answer for a wide range of complex trajectory queries. Their approach is inspired by ontologies,
but the resulting system they propose is database-based.

Apart from pure mining and knowledge discovery, there are also approaches that exploit trajectory semantics for different purposes. For example,
Richter et al.~\cite{RichterSL12} use geographical knowledge on POIs to compress trajectories while maintaining an acceptable information loss.
Monreale et al.~\cite{Monreale2010} discuss the privacy issues of semantic trajectories. Whenever a user trajectory crosses locations that may enable to infer sensitive information on the trajectory user, such as an hospital, a privacy issue arises. To solve such problem, they propose a privacy model for semantic trajectories, and an algorithm to preserve user privacy modifying the trajectory representation: in a safe trajectory, sensitive locations are abstracted along a place taxonomy to mask them, while preserving the trajectory semantics.

%Another way to infer trajectories is by ad-hoc devices and algorithms, like in \cite{DBLP:journals/gis/LaubeDP11}, where an algorithm for decentralized data mining on data from mobile geosensors is presented.While some research has focused on the efficient detection of movement patterns in centralized database systems (see, for example, \cite{laube04}), the work in \cite{DBLP:journals/gis/LaubeDP11} introduces a  new distributed algorithm for flock detection, and an empirical evaluation of the performances on simulated data and on real animal trajectories. \ec{dov'e' la semantica?}

%--------------------------------------------------
\subsection{Visualization of Semantic Trajectories}
Visual Analytics, together with Information Visualization, provides the instruments to empower human capacity for distillation and knowledge extraction from very large data repositories. In particular, Visual analytics develops intelligent visualization for data analysis.
%Going forward from pure visualization, Visual Analytics is a research area that develops intelligent visualizations for data analysis.
%It allows to synthesize information and derive insight from vast amount of data.%, sometimes even not internally coherent or noise-affected.The supported operations on the data range from the pure visualization, to their clusterization, from transformation of data segment for contextualization, to the analysis of presence, density and flow represented by the trajectories.
The research community in this area proposed several tools to improve the visualization of geographical data, bringing to the development of the area of GeoVisualization and Geo Visual Analytics. Not to be neglected is the contribution in stressing the contextual information attached over the trajectories, that allows its refinement and
classification \cite{andrienko2012visual}.

One of the main advantages of these visual techniques is the possibility to confirm expected patterns by detecting them, but also to observe the
emergence of unexpected ones. This can guide the users towards the revision,  either of the collection, extraction, distillation or representation
mechanism, or the model updating. Another observed effect is the possibility to improve the effectiveness in decision making process
by people: this can result from the availability of filtering, aggregating and drilling down functionalities in the visualisation interface.
% In fact, one of the problems that arise with the usage of geographical trajectories for the human consumption is the difficulty to extract
% knowledge from the raw data, normally extracted from raw events, well suited for the computer direct usage, and over-abundant in respect
% of the parsing capabilities of a human user.
% In fact, a part of the researches in the field of Information Visualization is devoted to provide more effective ways to represent the data,
%  in order to support the human judgment. This enables the detection of expected patterns or relationship and the emergence of unexpected ones,
% that can direct the user toward a revision of the collection, extraction, distillation or representation mechanism.
% In particular, it was found that the effectiveness for decision making by people can be enormously improved by providing capabilities for filtering
% and drilling down on relevant details starting from very high level aggregated information: this will be possible by founding together multidimensional
% visualizations with interaction capabilities \cite{robert_spence2007information}.

For the specific task of visualizing the Geo-Spatial data enriched with temporal information --which Semantic Trajectories is a subtype-- a
recent review from Andrienko et al. \cite{DBLP:journals/vlc/AndrienkoAG03} presents some possible techniques, working as a reference framework
for choosing the techniques that better fit the specific characteristics of the data to be represented and the objectives of the analysis.

Other works that address visualization to offer knowledge to the user are present in literature, such as the Weka-STPM
tool~\cite{Bogorny2011}. Beyond the pre-processing of data to semantically annotate trajectories and mining them, it includes also a visualization
interface for the semantic patterns extracted, such as frequent STOPs, MOVEs, and sequential STOPs. Another approach proposed by Bakshev et al.~\cite{BakshevSMVC11}
proposes a framework for trajectory visualisation and querying, where the semantic context of trajectories is modelled as an application domain ontology.

 In this area, the work of Andrienko and Andrienko is particularly relevant and a reference for the research community. In ~\cite{Andrienko:2011},
Andrienko et al. discuss how visualization and the graphical representation
of object movement can help understand its meaning, and present a conceptual framework about the possible types of information that can
be extracted from movement data. Currently the established visualization techniques for geographical data are {\em animated map} and {space-time cube}
(see, for example,
\cite{DBLP:journals/vlc/AndrienkoAG03}), which enhance understanding taking into account also the temporal dimension of data to support data analysis.

The space-time cube is also used by Zhong et al.~ \cite{Zhong2010} to design a method for semantic visualisation of trajectories based on the notion of events,
that are modelled as ADTs. Each event is characterised by the actor that does it, and by the place and the time it occurs. Moreover, levels of detail
(LOD) are associated to each event type.

Finally, \cite{Lau10} evaluates the importance of contextual information derived by geographical knowledge for visual analytics approaches to enhance
the understanding of human behaviour.
\subsection{Comparison with the proposed approach}
%----------------------------------------------------------
With respect to the current state of the art in Semantic Trajectory, our work has some distinguish characteristics and innovative aspects that we discuss in this section.  Referring to the previous classification of the research on this topic, the main contribution of this paper can be accounted to Knowledge Discovery, because we exploit semantically annotated trajectories for the discovery of movement patterns. However, our work addresses also the representation of trajectories and their semantics, therefore we compare it with the research in both areas.

In our approach, both trajectories and patterns are represented in an application domain ontology that extends a top-level ontology for representing moving objects.
Differently from work on trajectory segmentation that infers implicit semantics of episodes by processing the raw features of the trajectories or from the contextual knowledge, we adopt a reverse approach: taken spatio-temporal events with explicit semantics, we reconstruct the trajectories that describe the movements from one event to another.

In the test case scenario we propose, we start from Container Status Messages that encompass an explicit description of the activities that are undergoing on containers in a port, and from these labelled STOPs we reconstruct the container trajectories. The case of vessels is slightly different: we first aggregate container events to derive the implicit semantics of vessel events, and from them we build vessel trajectories as in the case of containers. However, we take into consideration the underlying geographical knowledge to distinguish among ports and other types of locations, that do not intervene in the patterns we discuss as examples.

Our approach has in common with \cite{Wannous2013,Baglioni08,BaglioniMRTW09,YanQuery08,BakshevSMVC11} the use of ontology for the representation of the domain and expert knowledge.
The usage of DL axioms for automatic reasoning on moving object data is applied in particular by \cite{Wannous2013,Baglioni08,BaglioniMRTW09}. Specifically,
similarly to Baglioni et al.~\cite{Baglioni08,BaglioniMRTW09}, we focus on the discovery of patterns expressed as ontology axioms and on the retrieval of ontology instances that verify such patterns.
However, even if the general approach is the same, with respect to \cite{Baglioni08,BaglioniMRTW09}, we go a step forward in term of complexity of domain knowledge and axioms.
In the application scenario we have considered for testing, the design of the MCO includes multiple moving objects (i.e., containers and vessels), and the ontology axioms formalizing anomalous patterns involve different semantic trajectories for these objects. In particular, usually more than one vessel is used to accomplish a container shipment: in transshipment operations, containers are unloaded from one vessel to another, and continue for another step of the trip. Transshipments can occur several times along a container trajectory. This implies that, to verify if a container trajectory is anomalous, we have to compare it with several vessel trajectories.

Moreover, differently from~\cite{Baglioni08,BaglioniMRTW09}, we translate axioms into DL queries, and evaluate according different implementation settings, considering combinations of different DL query languages and APIs and reasoning engines.
By contrast, Baglioni et al. tested their approach in~\cite{BaglioniMRTW09} importing the domain ontology in ORACLE and using OWLPrime to test the axioms.
In our case, we considered also this implementation alternative, but we discovered that OWLPrime is too limited to express the complexity of the axiom
conditions we have specified for the application case of maritime containers.

Our work has some similarities with~\cite{DBLP:journals/corr/abs-1205-1796}: actually, the authors have elaborated a meta-model to represent moving objects
using a mapping ontology for locations; despite this similarity, in extracting information from the instantiated model during the evaluation phase,
they seem to rely on a pure SQL-based approach, whether we rely on semantics queries.

%-----------------------------------------------------
\section{Description Logics (DL)} \label{sec:dls}
%-----------------------------------------------------

In this section we introduce the main features of DLs \cite{DBLP:conf/dlog/2003handbook}, that are the foundational basis of our formalization.
%., on which we rely to represent and reason on  the domain of maritime containers.
In DLs, the domain of interest is modeled by means of individuals,
concepts, and roles, denoting objects of the domain, unary predicates,
and binary predicates respectively.
Concepts correspond to classes, which are sets of objects,
while roles correspond to relations, i.e., binary relations on objects.

The basic syntactic building blocks of DLs are atomic concepts ($A$),
and atomic roles ($R$). Complex concepts (denoted by $C$ or $D$) can be built
from them inductively according to the syntax in the upper part of Table~\ref{tab:dl}.\\
%\noindent $C::=A~|~\top~|~\bot~|~\non C~|~C \sqcap D~|~C \sqcup D~|~C \sqcap D~|~\exists R.C~|~\forall R.C$\\
%\noindent In DLs, $\top$ is defined as universal concept, and $\bot$ is defined as bottom concept, such that $\bot=\non\top$.
From a semantic point of view, concepts are interpreted as subsets of an abstract domain, while roles are interpreted as binary relations over such a domain.
More precisely, an \emph{interpretation} $(\Delta^\Ical,\cdot^\Ical)$ consists of a domain of interpretation $\Delta^\Ical$, and an interpretation function
% $\cdot^\Ical$ mapping every atomic concept $A$ to a subset of $\Delta^\Ical$ and every atomic role $R$ to a subset of $\Delta^\Ical\times \Delta^\Ical$, as formalized by column {\bf Semantics} of Table~\ref{tab:dl}.
%The basic syntax and semantics of DLs is summarized in Table \ref{summary}.

%---------------------------------------------------------------------------------------------------------------------------------------------
\begin{table*}
\scriptsize
\label{summary}
\begin{center}
\begin{tabular}{|c|c|c|}
\hline
\textbf{Description} & \textbf{Syntax} & \textbf{Semantics} \\ \hline
universal concept & $\top$ & $\Delta^\Ical$ \\
bottom concept & $\bot$ & $\bot=\non\top$ \\
atomic concept & $A$ & $A^\Ical$ \\
concept negation & $\non C$ & $\Delta^\Ical\setminus C^\Ical$ \\
intersection & $C\sqcap D$ & $C^\Ical \cap D^\Ical$ \\
union & $C\sqcup D$ & $C^\Ical \cup D^\Ical$ \\
existential restriction & $\exists R.C$ & $\{x \in \Delta^\Ical~|~\exists y \in \Delta^\Ical,(x,y)\in r^\Ical \wedge y\in C^\Ical\}$ \\
universal restriction & $\forall R.C$ & $\{x \in \Delta^\Ical~|~\forall y \in \Delta^\Ical,(x,y)\in r^\Ical \rightarrow y\in C^\Ical\}$ \\
\hline
transitive role & $r_{T}$ & $(x,y)\in r_{T}^\Ical$ and $(y,z)\in r_{T}^\Ical$ imply $(x,z)\in r_{T}^\Ical$\\
nominal & $\{o\}$ & $\{o\}^\Ical$ \\
\hline
\end{tabular}
\end{center}
 \caption{Syntax and semantics of the DL features}\label{tab:dl} %\ec{aggiungere colonna per distinguere le due parti}}
\end{table*}
%---------------------------------------------------------------------------------------------------------------------------------------------

% \begin{table}
% \label{ex-sof}
% \begin{center}
% \begin{tabular}{|c|c|c|}
% \hline
% \textbf{Description} & \textbf{Syntax} & \textbf{Semantics} \\ \hline
% transitive role & $r_{T}$ & $(x,y)\in r_{T}^\Ical$ and $(y,z)\in r_{T}^\Ical$ imply that $(x,z)\in r_{T}^\Ical$\\
% %functional role & $r_{F}$ & $(x,y)\in r_{F}^\Ical$ and $(x,z)\in r_{F}^\Ical$ imply that $y=z$\\
% nominal & $\{o\}$ & $\{o\}^\Ical$ \\
% \hline
% \end{tabular}
% \end{center}
% \caption{Extra-properties to represent the domain of containerised transportation}
% \end{table}
% Table~\ref{ex-sof} summarizes the extra-features added to $\mathcal{ALC}$ for the representation of the domain
% of containerized transportation.

A \emph{Knowledge Base} (KB) comprises two components:
the {\em TBox} and the {\em ABox}. The TBox is a finite set of terminological
axioms which make statements about how concepts are related to each other.
Generally, they have two forms: $C\equiv D$ or $C \sqsubseteq D$, where $C, D$ are concepts. The
first kind is called \emph{equalities} which states that $C^\Ical$ is equivalent to $D^\Ical$, and the
second is called \emph{inclusions} which states that $C^\Ical$ is a subset of $D^\Ical$ for all $\Ical$. %Since
%the \ALC's terminology is acyclic, we will focus on the acyclic terminology in
%the following.
The ABox is a finite set of individual assertions, which can be of two types: $C(a)$ or r$(a,b)$, where $C$ is a {\em concept}, $r$ is a {\em role},
$a,b$ are individuals. The first kind is called \emph{concept assertions} which states that
$a^\Ical \in C^\Ical$, and the second is called \emph{role assertions} which states that $(a^\Ical,b^\Ical)\in r$ for all $\Ical$.

The basic reasoning services in DLs are \emph{satisfiability} and \emph{subsumption}.
A concept $C$ is satisfiable in a $KB$ $K$ if $K$ admits a model in which the extension of $C$, i.e., the set of individuals that belong to $C$, is non empty.
By contrast,  $C$ subsumes $D$ in $K$ if $C^\Ical \subseteq D^\Ical$ for every interpretation $\Ical$ of $K$.
Subsumption can be easily reduced to satisfiability as follows: A concept $C$ is subsumed by a concept
$D$ in $K$ if and only if $C \sqcap \non D$ is not satisfiable in $K$. Upon that it is sufficient to consider concept satisfiability only.

We refer to the DL $\mathcal{ALC}$\cite{DBLP:conf/dlog/2003handbook} to represent and reason
on the domain according to its features. Moreover, we have extended its expressivity to represent the domain of containerised transportation.
In particular, \emph{nominals} and \emph{transitive} roles are needed in this context.
Nominals are necessary to identify the locations involved in a suspicious pattern.
Transitive roles are necessary to bind every container event with all the subsequent ones.
The addition of these two features does not influence the complexity of the basic reasoning services, which, in presence of an acyclic
TBox\footnote{a TBox is acyclic iff no concept name uses itself.},
remains PSpace-complete as in $\mathcal{ALC}$ \cite{DBLP:conf/dlog/2003handbook}. Although the reasoning is of a relatively high
complexity, the pathological cases that lead to the worst case complexity rarely occur in practice \cite{DBLP:conf/dlog/2003handbook}.

% \ec{aggiungiamo che il worst case scenario e' raro, con citazione?}
%\pv{Essenziali per noi sono i ruoli transitivi e i nominali, cosa che ci porta alla PSpace completezza (e il worst case e' raro secondo Baader e soci). Questo tenendo conto che nella parte implementata non abbiamo gerarchie di ruoli, che ci riporterebbero a ExpTime... Togliendo i ruoli funzionali e gerachie di ruoli sarebbe corretto dire che abbiamo usato un sottoinsieme di SHOIN, cioe' SO...anche se SO e' fin  troppo grande, perche' non non usiamo concetti negati}
% In this context, functional roles are not necessary, because the procedure for the population of the ontology
% guarantees that a container itinerary has only one start and one arrival location.

%===================================================================================
\section{A Methodology for Trajectory Pattern Discovery}
\label{sec:methodology}
%===================================================================================

In this section we present the methodology we propose for the discovery of patterns and behaviors in moving object trajectories. Specifically,
given a dataset of moving object trajectories, we want to retrieve the trajectories that follow a given pattern, i.e., have a certain movement behaviour.
Our approach strongly relies on ontology and on the DL formalism:  we use ontology for the representation of the moving object application domain, and DL axioms for the specification of
the patterns.

In the following, we define the graphical formalism we use in the paper for describing the ontology design;  then, using such formalism, we introduce
a top-level ontology for modelling moving object trajectories, namely the Moving Object Ontology (MOO).
Afterwards, we discuss how the MOO can be extended to formalize the semantics of a specific application domain,
and explain how trajectory patterns can be formally defined to enable instance retrieval.
Finally, we describe the implementation workflow we have developed for itinerary pattern discovery.

%--------------------------------
\subsection{Ontology diagrams}
%--------------------------------
In the paper we introduce the ontology design we apply through the support of ontology diagrams describing the {\em concepts}
and the {\em roles} between them, where concept and role have the semantics we have introduced in Section~\ref{sec:dls}.
An example of ontology diagram is given in Fig.~\ref{fig:toponto}, that illustrates the MOO design.
%Roles are binary directed relations from one concept to another, have a name and their domains and co-domain can be restricted by means of ontology axioms.
We represent concepts as rectangles with rounded corners, while we depict roles as directed arrows.
For the sake of clarity, we do not report the concept's structural properties but describe them in the text whenever necessary.
In the text, the ontology names are emphasized (e.g., {\em Moving Object}).
However, within the discourse entity and concept names are used interchangeably where no ambiguity arises.

Concept {\em generalizations} are depicted as straight lines that go from low-level to top-level concepts, similarly to the IS-A relation of object-oriented models.
Starred labels ({\sf label*}) model one to many relationships. %Arrows with double heads represent a role and its inverse.
Underlined arrow labels represent roles that have been re-defined in sub-concepts; the corresponding domain and co-domain are restricted accordingly by means of ontology axioms.
%\ec{Hierarchies of roles are used in the ontology design, but they are not actually implemented  in OWL to further reduce the computational complexity of the representation.}

%----------------------------------------
\subsection{Moving Object Ontology (MOO)}
\label{subsec:moo}
%----------------------------------------
%-------------------------------------------------------------------------------------------------------------
\begin{figure*}[t]
\begin{center}
  \includegraphics[width=15cm]{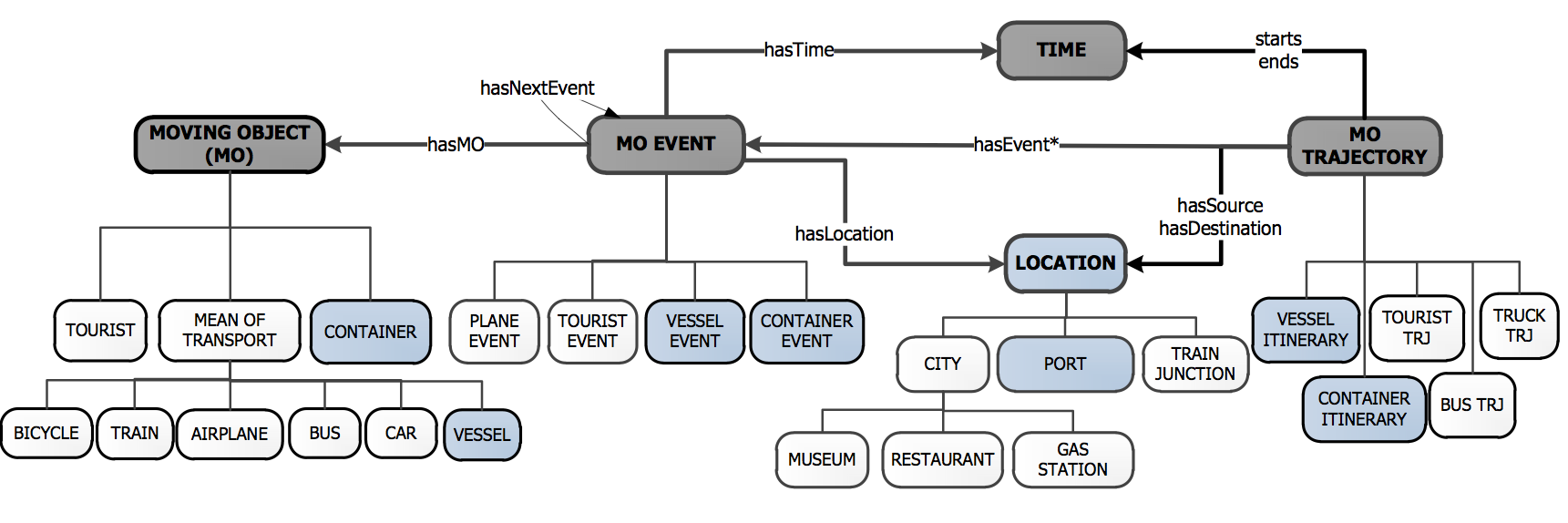}
\end{center}
\caption{Moving Object Ontology, a top-level ontology for moving object trajectories \label{fig:toponto}}
\end{figure*}
%--------------------------------------------------------------------------------------------------------------
The fundamental entities of the MOO  abstract the features that are common to different domains focusing on the movement of some kind of object, such as traffic analysis for route planning, pedestrian trajectory analysis, animal movement analysis, detection of shipping corridors for maritime surveillance, etc.

The concepts formalising these entities are depicted in Fig.~\ref{fig:toponto}, namely, {\em Moving Object} (MO), {\em MO Trajectory}, {\em Location}, {\em Time}, and {\em MO Event}.
{\em MO} formalises any class of objects that move, such as cars, persons, airplanes, buses, etc.
{\em MO Itinerary} models the semantically enriched movement of the MO, defined as {\em sequences} of {\em MO Event}s.
%An event formalizes something that happens to the moving object at a specific {\em Time} in a particular {\em Location}, such as a city, a train station, a museum.
Events are crucial concepts in our modelling, because we rely on them to %define the object movement and to
leverage the trajectory %movement
semantics.
Events describe the activities accomplished by the MO, each occurring at a specific {\em Time} in a particular {\em Location}. For example, a container in a port is {\em loaded} on a cargo vessel; a car at a gas station is {\em refuelling}.

Event semantics can either be {\em explicit}, i.e., declared in the data, as we see for the case of containerized transportation, or {\em implicit}, but nevertheless inferrable from other contextual information: for example, knowing that a person is in a restaurant at lunch time we can likely infer that this person is eating.
%Implicit event semantics has been used for example in \cite{BaglioniMRTW09} to analyze touristic itineraries.
Event semantics may also help infer additional information on the object activity: for example, after a container has being loaded on a vessel, we can foresee that it will start soon travelling.

%Since events are timestampted, for each  moving object we can obtain its {\em event sequence}; navigating the sequence, we can follow what the MO has done along its trajectory, i.e., the sequence of locations the MO passes through.
We can navigate the events in an itinerary according to the sequence they occur, relying on their timestamps. Navigating the sequence, we can follow the MO along its trajectory and along the activities it has done during the itinerary.
Moreover, event sequences are also modelled intensionally in the MOO through the {\em transitive} property {\em hasNextEvent}, which links each event to the next event in the sequence.

In Fig.~\ref{fig:toponto} we have depicted also the roles between concepts. For example, events are connected to {\em MO} by the role {\em hasMO}; by role {\em hasLocation} to
{\em Location}, which generalizes {\em City}, {\em Port}, {\em Train Junction}, etc; and by role {\em hasTime} to  {\em Time}.

 %Let's suppose that for moving object $o$ a temporally ordered sequence of events $\cal S {e_t| t = 1 \ldots s}$ have been stored in the ontology. Givent two events in the sequence $e_i$ and $e_{i+n}$, happening at time $i$ and $i+n$, with $n \gt 0$, refer to two different locations $l_i$ and $l_{i+n}$, with $l_i \neq l_{i+n}$, we infer that $o$ has moved from $l_i$ to $l_{i+n}$ sometime between $i$ and $i+n$.

%----------------------------------------
\subsection{Domain Ontology and Patterns}
%----------------------------------------
To model the entities of the application domain of interest, ontology concepts and roles in the MOO have to be extended.
For example, in Fig.~\ref{fig:toponto} we have extended the concept {\em Moving Object} to represent {\em Car}s, {\em Person}s, {\em Airplane}s, {\em Bus}es.
In the next section, we see how the MOO has been extended to model the domain of containerized transportation.

In our application scenario, we are interested in formalizing movement patterns and in retrieving the trajectories that comply with the behaviour such patterns express.
Patterns may be specified directly in the domain ontology as axioms. An axiom defines, using the DL syntax, a new class of objects, whose ontology instances are those verifying the axiom conditions.
Therefore, to retrieve the trajectory instances that verify the patterns, it is sufficient to check the pattern axioms against the ontology.

As an alternative, axioms can be transformed into explicit DL queries, which can be used to query the ontology instances. This solution
enlarges the implementation possibilities because different languages and APIs
are available to express them. Currently, the most used ones are OWL-API~\cite{Horridge2011-owlapi} and SPARQL-DL~\cite{Sirin07sparqldl}, that we have tested in the experimental
evaluation in Section~\ref{sec:exp}.

%----------------------------------------
\subsection{Pattern Discovery Workflow}
%----------------------------------------

\begin{figure*}[t]
\begin{center}
  \includegraphics[width=14cm]{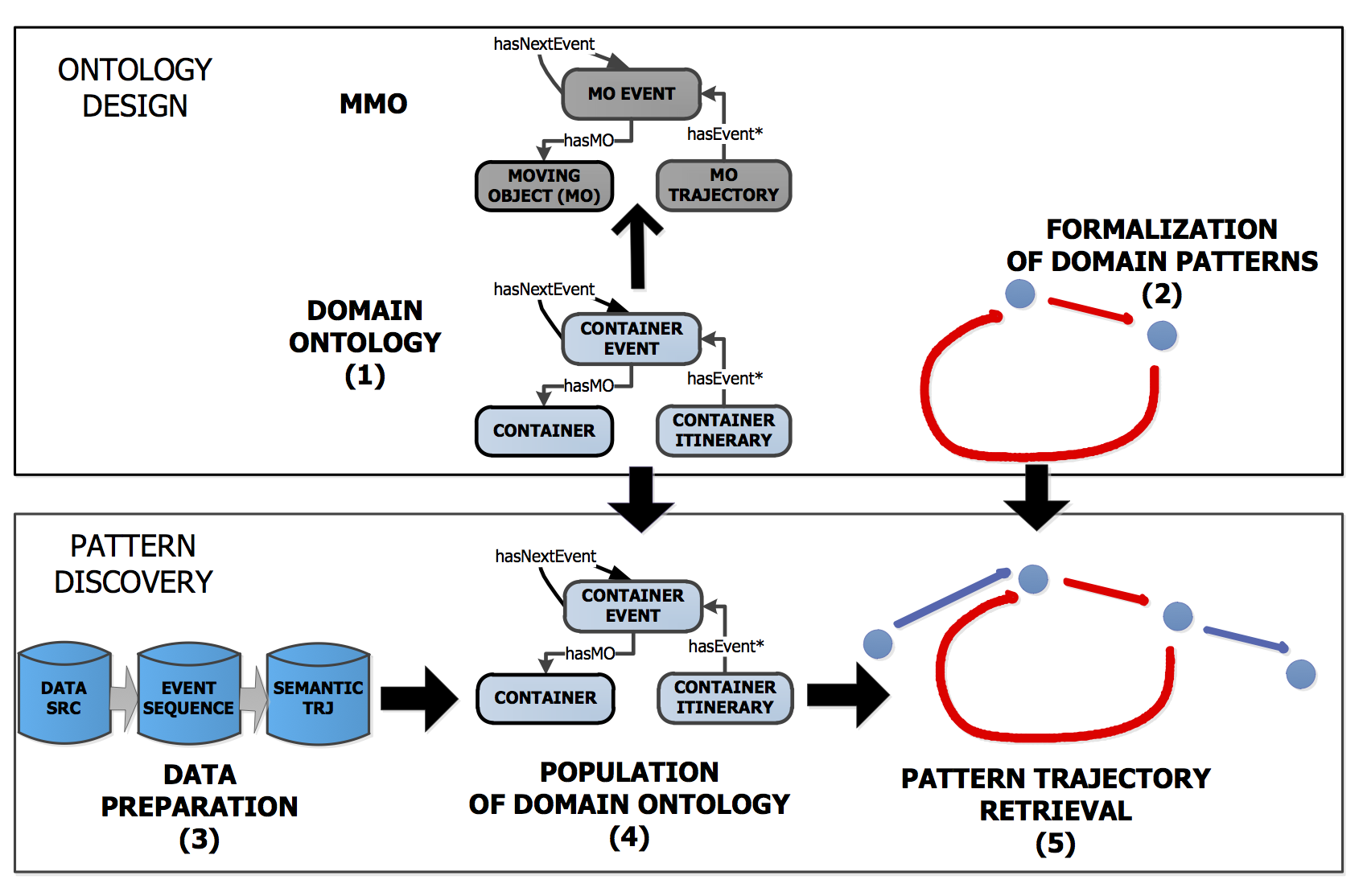}
\end{center}
\caption{Pattern Discovery with Ontology} \label{fig:process}
\end{figure*}

The complete workflow for pattern discovery is illustrated in Fig.~\ref{fig:process}.
Once the MOO is extended at step (1) to model the application domain
and (2) the movement patterns have been defined as described above,
we can proceed with the development of the pattern discovery tool. At step (3), data have to be selected, to extract the event sequences, and
the event semantics must be made explicit, annotating the moving object trajectories.
% At step (4), the domain ontology is populated, transforming the annotated trajectories in ABox instances. On the populated ontology we can reason and infer new knowledge, retrieving the trajectory patterns: we can check the ontology axioms, or run the equivalent DL queries.

%%------------------------------------------------------------------------------------------------------------------------------------------------------------------------
%------------------------------------------------------------------------------------------------------------------------------------------------------------------------

%
%===================================================================================
\section{Maritime Container Ontology} \label{sec:mco}
%===================================================================================

In \cite{DBLP:conf/geos/VillaC11}, we proposed the Maritime Container Ontology (MCO) to represent the domain of the maritime containers.

In the remaining of the section,  we describe the MCO design, that extends the MOO formalised above to define containers,  %and shipments; further on, we focus on the formalization of
container and vessel itineraries, leveraging on the semantics of events.
Herein we do not report the detailed design of shipments and shipment phases, that goes beyond the scope of the paper. We refer the interested reader to  \cite{DBLP:conf/geos/VillaC11} for the details.

In the ontology diagrams in the section, we use the following convention for role inheritance: roles in {\it italic} are inherited by the MOO as they are, while roles whose name is \underline{underlined} are inherited roles that have been specialized to refer to specific sub-concepts.
\subsection{Containers and Shipments}\label{sec:container}
%-------------------------------------------------

%--------------------------------------------------------
\begin{figure*}[t]
\begin{center}
%\framebox{
		\includegraphics[width=15cm]{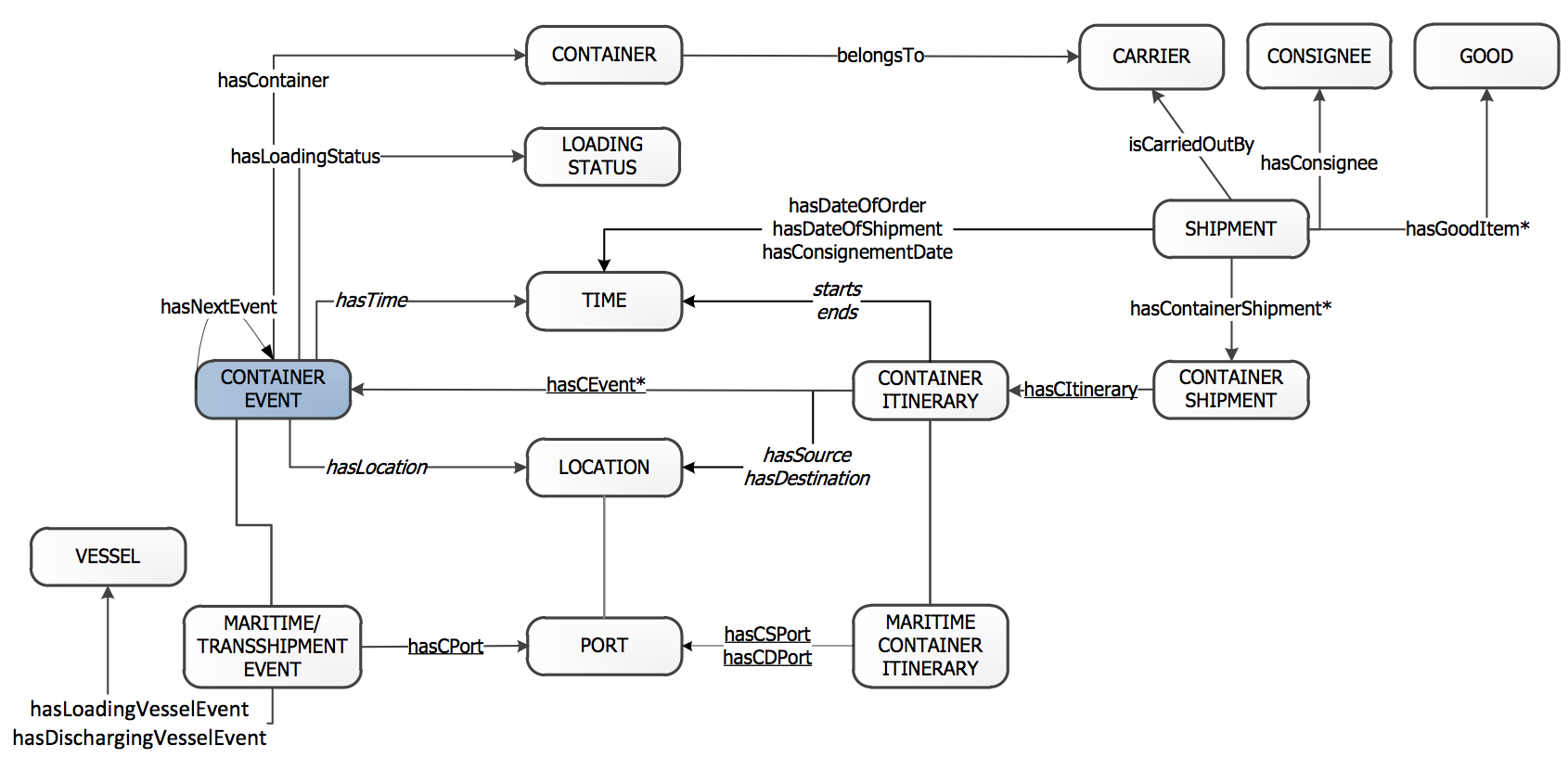}
%		}
	\end{center}
\caption{{\em Container Event}, {\em Container Itinerary} and {\em Shipment}}\label{fig:container}
\end{figure*}
%--------------------------------------------------------

In the MCO every container is modelled by an instance of the concept {\em Container}, which extends {\em Moving Object} in the MOO (see Fig.~\ref{fig:toponto}). Each container has a unique identifier, that maps an ISO 6346 \cite{ISO6346} identification code, i.e., the BIC code \footnote{BIC codes are assigned by the {\em Bureau International des Containers et du Transport Intermodal} (BIC).}. %, according to the International Organization for Standardization (ISO) 6346 standard \cite{ISO6346}.
Every container belongs to a {\em Carrier}, i.e., a shipping or a leasing company, which leases the container to a carrier, to whom it is connected by the role {\em belongsTo} (see Fig.~\ref{fig:container}).

Each {\em Shipment} is handled by a {\em Carrier} to deliver a set of {\em Goods} and encompasses the dates when the order has been placed, shipped and delivered to a {\em Consignee}. A shipment is made by at least one {\em Container Shipment}; each {\em Container Shipment} refers to a single container and has one {\em Container Itinerary}.

\subsection{Container Itineraries and Events}
%----------------------------------------------------------
A {\em Container Itinerary} is defined by all the events occurring to a container to accomplish a shipment. These encompass the transport, which is mainly performed by sea, but also the operations to prepare and conclude the shipment. Therefore, a container itinerary goes beyond the mere trajectory of the container, and represents the complete history of the shipment performed using the container.

%\ec{aggiungere pezzo sui marittime itinerari}

A {\em Container Event} describes any deed undertaken on a container, such as {\em Loaded to vessel}, {\em Discharged at port}.
%; pairs of container events define a shipment phase.
{\em Container Event} extends {\em MO Event} in the MOO and  refers to the {\em Time} it occurs (e.g., 26th of November 2020) and the {\em Location} where this event took place. This can be either a port in intra-customs transport, or a train station or a city in inland transportation.
%\ec{Note on timestamps: Time of reference is not necessarily the same, i.e., validity time or transaction time; Granularity is {\em minutes}}

Each container event refers also to other information dimensions, including the container {\em Loading Status} (i.e., empty, full) and, for events referring to transportation, to a {\em Mean of Transport}, in particular {\em Vessel}s for {\em Maritime Container Event}s which are the events occurring during the maritime transportation.

%--------------------------------------------------------
\begin{figure*}[t]
\begin{center}
%\framebox{
		\includegraphics[width=16.5cm]{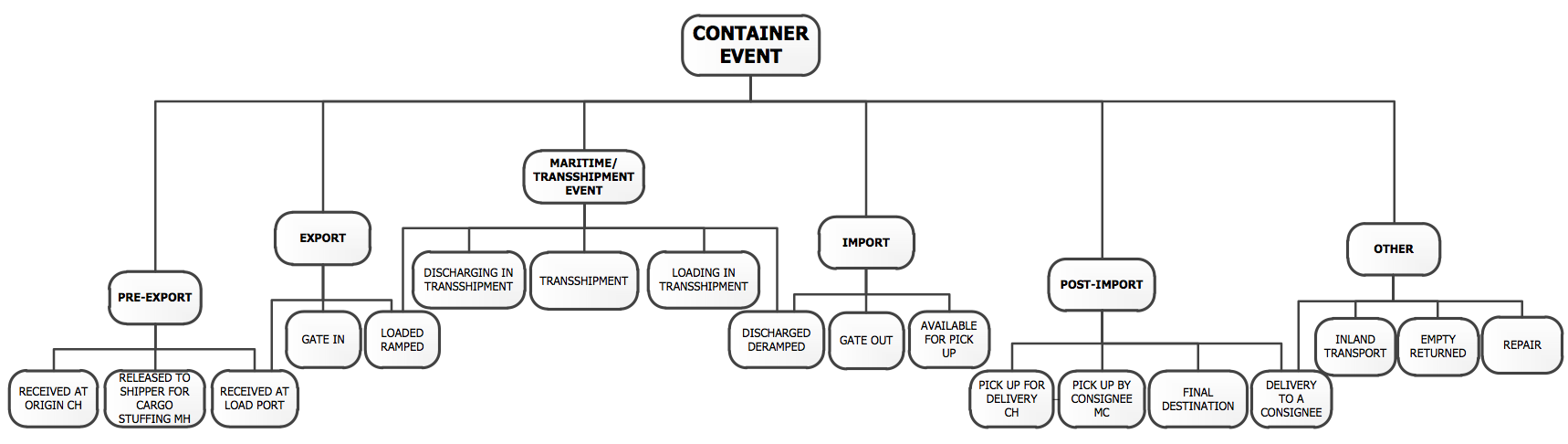}
%		}
	\end{center}
\caption{Concepts for reference container events. These events are used to classify the events specified by the carriers. \label{fig:events}}
\end{figure*}
%--------------------------------------------------------

There is no standard for event descriptions, and each carrier adopts a different one.
Within the project an effort towards standardization of container events has been promoted, and the outcome has been formalized in the MCO:
in Fig.~\ref{fig:events} we report eighteen events, classified among four classes of top-level events: {\em Trip Start}, {\em Maritime/Transshipment Event}, {\em Trip End}, and {\em Other}.
Each event, as specified by the carrier, is mapped to an instance of one of the concepts specified in the figure. This mapping simplifies the representation of the application domain, and enables to abstract from the contextual knowledge of the carrier vocabulary when defining the axioms for anomalous patterns, as we will see in Section~\ref{sec:axioms}.
%Such events have a correspondance with top-level shipping phases.

Top-level events characterize the different phases of a shipment.
%and have a correspondance with the top-level shipping phases described above.
In Fig~\ref{fig:container}, only {\em Maritime/Transshipment Event}s are shown to focus on the main events occurring during the maritime part of a container itinerary, that is loading to and discharging from vessels during the maritime transportation. For such events, the {\em Vessel} the container has been loaded to or from which it has been discharged is also reported (roles {\em hasDischargingVessel} and {\em hasLoadingVessel}). In case a transshipment occurs in an intermediate port, the vessels involved are always two and the two roles are filled in accordingly.
We can see in Section~\ref{sec:axioms} that transshipments from one vessel to another play an important role in defining suspicious patterns.

Other events, such as {\em Released to Shipper for Cargo Stuffing} and {\em Empty Returned}, do not describe any container movement, but deeds occurring to prepare the container for the shipping at the source port or to complete it at the port of destination.
They may be helpful to confirm the presence of a container in a port at the begin and at the end of a shipment, as well as to define the temporal period spent by the container in a port, helping characterise the itinerary with better accuracy. % and are therefore included in the container itinerary.%, which does not describe only the physical movement of the container but its complete history during a particular shipment.

\subsection{Vessels Events and Itineraries}\label{subsec:vesselRoutes}
%-----------------------------------------------
%Instances of {\em Means of Transport} represent trucks, trains, cargo ships and any other veichle used for carrying goods in containers.
In the MCO, we focus in particular on cargo vessels, because most of the import-export of goods is performed by sea.
Vessels in the ontology are uniquely identified through their name and, when available, the International Maritime Organization (IMO) number.

%--------------------------------------------------------
\begin{figure}[b]
\begin{center}
%\framebox{
		\includegraphics[width=8cm]{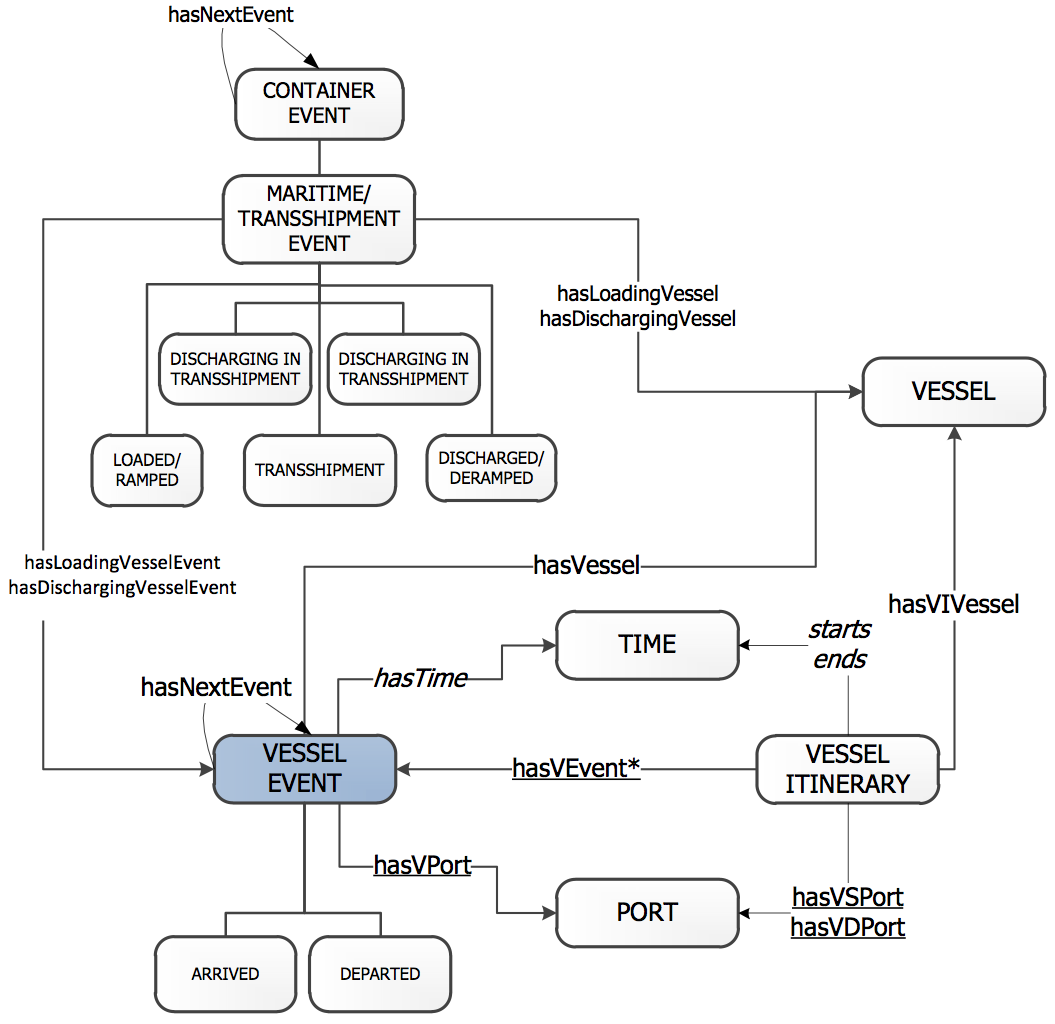}
%		}
	\end{center}
\caption{{\em Vessel Event} and {\em Vessel Itinerary}}\label{fig:vessel}
\end{figure}
%--------------------------------------------------------

We focus on {\em Arrival} and {\em Departure} events (see~\ref{fig:vessel}), that occur in {\em Port}s and are sufficient to define the vessel movement.
%Each one-step trip between two ports is represented by an instance of {\em Vessel Itinerary}. \ec{e' vero? ma poi alla fine non prendiamo sempre in considerazione tutta la sequenza della vessel? Non sarebbe meglio dire quanto segue?}\pv{secondo me si}
A {\em Vessel Itinerary}, as above, models extensively a sequence of events, which is also defined intensively through the transitive relationships {\em hasNextEvent} inherited by {\em Moving Object Event}. As before, instances of {\em Vessel Event} model the STOPs of a {\em Vessel Itinerary} \cite{Bogorny2010,Spaccapietra}.
In particular, as described above, a {\em Transshipment} of a container involves two different vessels.
%, therefore it is related with at least two different vessel events (a {\em Vessel Discharging} and a {\em Vessel Loading}).
%As we will see in Section~\ref{sec:axioms}, this implies also that, to check the consistency of a container itinerary that involves a transshipment, we have to compare it with two distinct {\em Vessel Itineraries}s. \footnote{In the generic case, in which the container has been transshipped several times, we have to consider multiple vessel itineraries.  Let the number of transshipments done on the container be $n$; then, the vessel itineraries involved are $n+1$.}

%=====================================
\section{Suspicious Patterns} \label{sec:axioms}
%=====================================

On top of the semantic model formalising the domain knowledge, we developed the axioms for the discovery of anomalous patterns.
In particular, here we present two suspicious patterns: namely, {\em \cycle} and {\em \trans}. Such patterns have been defined
in a collaboration with experts of Custom's Risk Intelligent Department, and are patterns that potentially suggest
some fraud activity has occurred, because they carry out unnecessary operations that entail extra costs or delays for the shipper.

% The first pattern describes a {\em loop} occurring in a container itinerary, i.e., the container is loaded on a vessel in a port $P$; afterwards, in an intermediate port, the container is transshipped to another vessel, that comes back to the port $P$ before reaching the shipment destination.
%
% The second pattern holds when {\em unnecessary transshipments} take place, i.e., the container is transshipped from its original vessel ($A$) on another vessel ($B$), in an intermediate port.
% Comparing the vessels routes, we infer that vessel $A$ and vessel $B$ go to the same
% destination. Such a transshipment appears unnecessary from a logistic point of view; however, whenever the container follows it, it may appear to be coming from the starting port of vessel $B$.

These patterns are  defined in the MCO as DL axioms.
%These two axioms are expressed in the description logic $\mathcal{ALCO}$ \cite{DBLP:conf/dlog/2003handbook}, that is $\mathcal{ALC}$ with nominals.
%As we will see, nominals are necessary in order to identify separately each involved location.
Each axiom combines ontology concepts with logical operators, defining implicitly the class of objects describing the
container itineraries following the corresponding suspicious pattern.

Both the axioms crosscheck container and vessel itineraries: this is because
cargo vessels transport thousand of containers during their trips, and usually pass through more than one port for each voyage.
For logistic reasons, when a vessel arrives in a port, some containers are {\em transshipped} %discharged and loaded to another vessel (namely, containers are {\em transshipped})
to reach the next port in their itinerary; at the same time, other containers are loaded to the vessel, that will continue its trip.
A container may be transshipped several times before reaching its destination, therefore, vessels routes do not coincide with
maritime container itineraries, but partially overlap with them.  To discover anomalies, we have to crosscheck container itineraries with
vessel trips, in order to discover the real trajectory followed by a container.

Suspicious pattern go beyond the simple patterns presented in similar approaches \cite{Baglioni08}, in particular because they involve multiple itineraries and events classes, i.e., each axiom evaluates a container itinerary and the itineraries of the vessels used for its shipment.
%This is because each container shipping is usually accomplished by more than one vessel, to which the container is transshipped during the trip.
%Moreover, Note also that each transshipment may involve several loading and unloading operations.
This is necessary because the container itinerary is not completely specified by its own, but to fully understand it we have to take into account loading and discharging operations and intersecting the container trajectory with those of the vessels used for its transportation.
Moreover, the semantics of the container STOPs~\cite{Spaccapietra} is not inferred from the place classification,
but is derived from event descriptions.

%Such patterns have been tested against the ConTraffic database~\cite{ctrf}. The experimental evaluation is described in Section~\ref{sec:exp}.

%=====================================
\subsection{Loop}%\label{subsec:cycle}
%=====================================

\begin{figure}[th]
\begin{center}
\includegraphics[width=8cm]{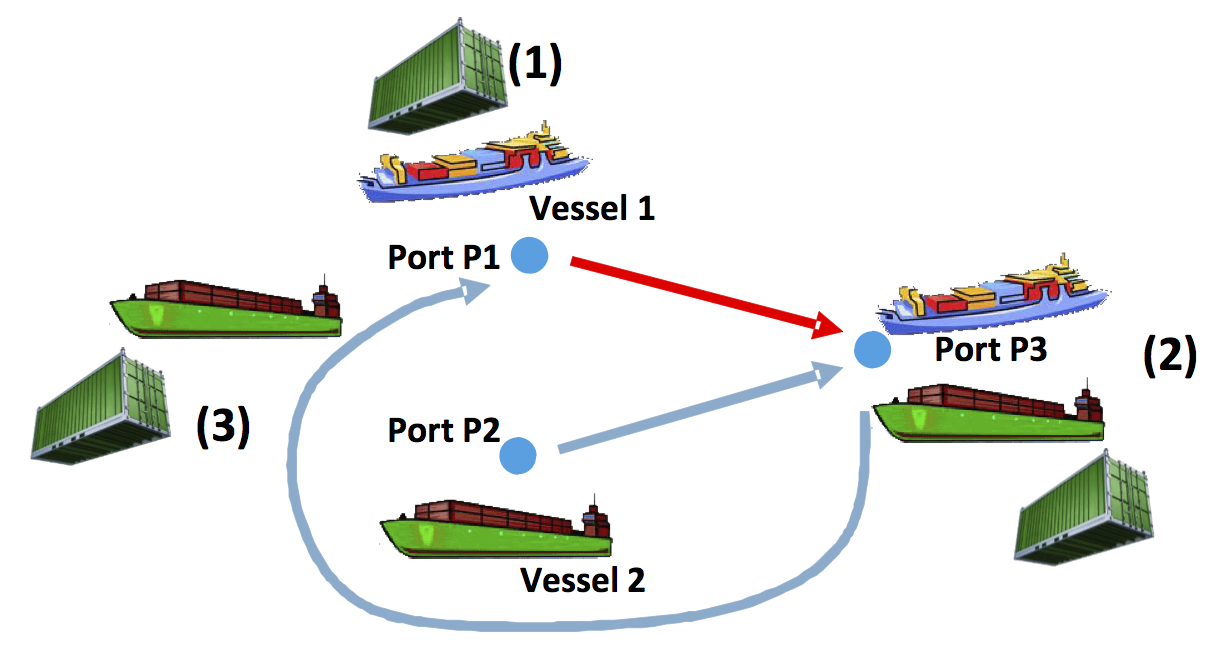}
\end{center}
\caption{Pattern {\em Loop}: (1) the container is loaded on $Vessel_1$ in port $P_1$;
(2) the container is transshipped on $Vessel_2$ in port $P_3$;
(3) the container is back in port $P_1$ before reaching its final destination} \label{fig:cycle}
\end{figure}

The pattern \cycle~is graphically depicted in Fig.~\ref{fig:cycle}.
A container is loaded on $Vessel_1$ in port $P_1$ at time $t_1$, with destination $P_x$.
At time $t_3$ $Vessel_1$ reaches the intermediate port $P_3$, where the container is transshipped on
$Vessel_2$.
Afterwards, $Vessel_1$ continues its itinerary, while $Vessel_2$ comes back to port $P_1$ before
reaching $P_x$.

%--------------------------------------------------------
%--------------------------------------------------------

Given the formalisation represented in Fig.~\ref{fig:container} and Fig.~\ref{fig:vessel},
the axiom that formalises pattern~\cycle~defines the class of container itineraries that involve a transshipment on a vessel that
comes back to port $P_1$ before reaching port $P_X$, as depicted in Fig.~\ref{fig:cycle}.
The corresponding DL specification is as follows:

%--------------------------------------------------------
\begin{patternImpl}{\em (axiom \cycle)}\\
\scriptsize
\begin{eqnarray*}
    &\lf{LoopP1\_P2}\equiv&
    \lf{MaritimeContainerItinerary}\\
    & &
    \sqcap \exists\lf{hasCISourcePort}.\lf{\{P1\}}\sqcap\\
    & &
    \exists\lf{hasCIDestinationPort}.\lf{\{PX\}}\sqcap\\
    & &
    \exists\lf{hasContainerEvent}.(\lf{Transshipment\_Event}\sqcap\\
    & &
    \exists\lf{hasLoadingVesselEvent}.(\exists\lf{hasNextEvent}\\
    & &
    .(\exists\lf{hasVPort}.\lf{\{P1\}}\sqcap\\
    & &
    \exists\lf{hasNextEvent}.\exists\lf{hasVPort}.\lf{\{PX\}}))))\\
\end{eqnarray*}
 \finepatt
\end{patternImpl}
%--------------------------------------------------------

The core of the axiom is the concept {\em Transhippment Event}, which allows to abstract from the specific definitions of transhippment
to avoid depending on different ways to describe the same events,
\hec{such as: aggiungere esempi di eventi che sono stati astratti}
combined with the role {\em hasLoadingVesselEvent} (see Fig.~\ref{fig:vessel}), which
links the container itinerary to the route of any vessel used for its transportation. The axiom \cycle~matches all the itineraries
in which a loading vessel comes back to the port of origin of a container
before reaching the shipment destination.
\hec{solo l'origine? un porto intermedio no?}

Note that it matches all cycle patterns, disregarding the number or transshipments done during the itinerary of the  container.
However, to be sure of pruning false positive cases, we have to take into account two dates; the first one is the container
arrival, and the second one is the arrive of the vessel that performs the loop: if they are in the same day, or in very close days,
we can be sure that the itinerary is suspicious; if they differ of months, of even years, then we can be in presence of a gap in the container
or vessel event sequence.

% The SPARQL-DL~\cite{Sirin07sparqldl} implementation described in Section~\ref{sec:exp}
% takes into account these two dates for every suspicious itinerary: in this way, it is sufficient to crosscheck them in order to filter false
% positive cases.

$P_1$ and $P_X$ are two nominal concepts that indicate two different ports.
%
%\hec{nota esplicita sul fatto che non si possono usare variabili nel linguaggio} To manage this in the implementation phase
%
To process all the ports in a dataset, the implementation described in Section~\ref{sec:exp} process the axiom iteratively
on all possible pairs of locations.

We also propose a slightly different specification of the axiom to describe the event that $P_1$ is not the
starting port for a container itinerary, but is one of the intermediate ports that the container reaches before arriving to the final destination.
In this case, we have to test the axiom considering for $P_1$ all possible values that come before $P_X$ in the trip.
The corresponding DL specification is as follows:

%--------------------------------------------------------
\begin{patternImpl}{\em (axiom \cycle~- intermediate ports)}\\
\scriptsize
\begin{eqnarray*}
    &\lf{LoopP1\_P2}\equiv&
    \lf{MaritimeContainerItinerary}\\
    & &
    \sqcap \exists\lf{hasCIEvent}.(\exists\lf{hasLocation}.\lf{\{P1\}})\sqcap\\
    & &
    \exists\lf{hasCIDestinationPort}.\lf{\{PX\}}\sqcap\\
    & &
    \exists\lf{hasContainerEvent}.(\lf{Transshipment\_Event}\sqcap\\
    & &
    \exists\lf{hasLoadingVesselEvent}.(\exists\lf{hasNextEvent}\\
    & &
    .(\exists\lf{hasVPort}.\lf{\{P1\}}\sqcap\\
    & &
    \exists\lf{hasNextEvent}.\exists\lf{hasVPort}.\lf{\{PX\}}))))\\
\end{eqnarray*}
 \finepatt
\end{patternImpl}
%--------------------------------------------------------

%The management of this situation is explained in Section \ref{sec:exp}.
%

%=====================================
\subsection{Unnecessary Transshipment}
%\label{subsec:portpat}
%=====================================

\begin{figure}[b]
\begin{center}
\includegraphics[width=8cm]{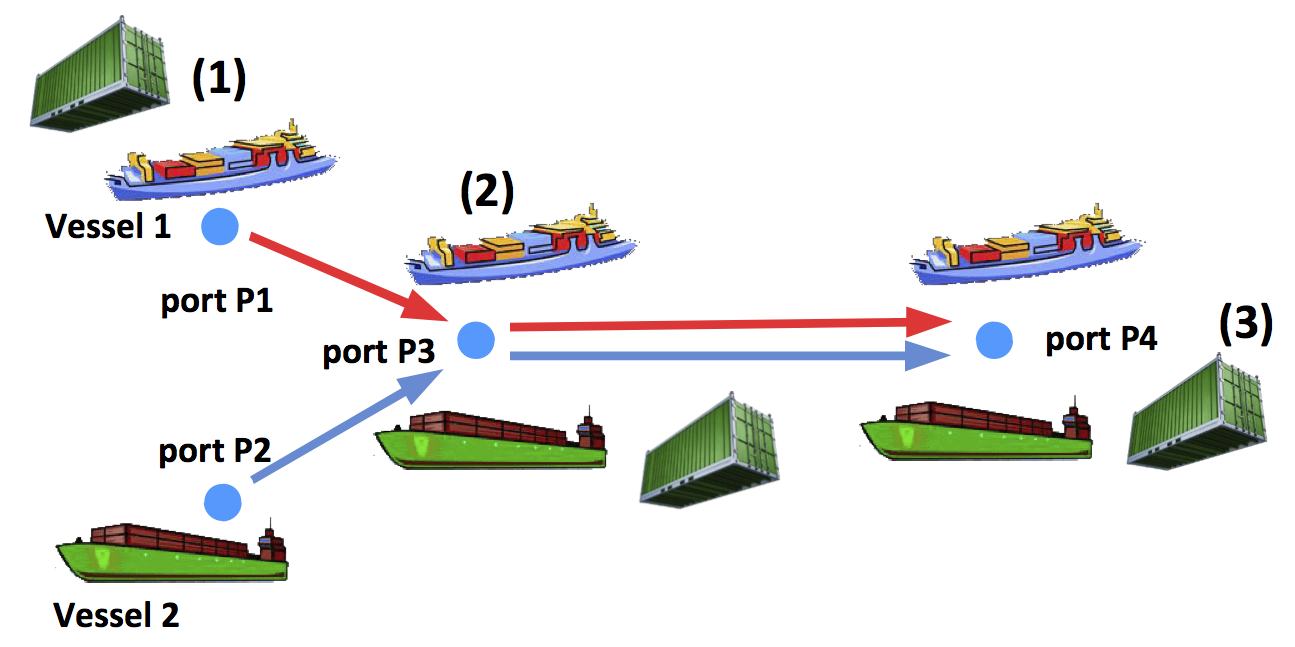}
\end{center}
\caption{Suspicious Pattern {\em Unnecessary transshipment}:
(1) the container is loaded on $Vessel_1$ in port $P_1$;
(2) the container is transshipped on $Vessel_2$ in port $P_3$;
(3) the container arrives at port $P_4$; also $Vessel_1$ reaches the same port}\label{fig:pattern2}
\end{figure}

Pattern \trans~is in Fig.~\ref{fig:pattern2}, where a container, loaded at time $t_1$ on $Vessel_1$ in port $P_1$, is
transshipped on $Vessel_2$ in an intermediate port $P_3$ at time $t_3$, and afterwards, both $Vessel_1$ and $Vessel_2$
arrive at port $P_4$, which is the container destination, therefore the transhippment was not necessary.
Such a manipulation in the container itineraries is often put in place to conceal the real origin of a shipment,
to take advantage of convenient duties agreement between the countries involved:
Indeed, thanks to such unnecessary transshipment,
a fraudulent shipper can easily manipulate the container documents pretending
that the shipment originated from the starting port of $Vessel_2$, i.e., port $P_2$, instead of $P_1$.
%
%--------------------------------------------------------
%--------------------------------------------------------

Given the formalisation represented in  Fig.~\ref{fig:container} and Fig.~\ref{fig:vessel}, the DL axiom formalizing pattern \trans~is as follows:
%--------------------------------------------------------
\begin{patternImpl}{\em (axiom \trans)}\\
\scriptsize
\begin{eqnarray*}
    &\lf{Unnecess\_TransP}\equiv&
    \lf{MaritimeContainerItinerary}\sqcap\\
    & &
    \exists\lf{hasCIDestinationPort}.\lf{\{P\}}\sqcap\\
    & &
    \exists\lf{hasContainerEvent}.(\lf{Transshipment\_Event}\sqcap\\
    & &
    \exists\lf{hasDischargingVesselEvent}.( \exists\lf{hasNextEvent}\\
    & &
    .(\exists\lf{hasVPort}.\lf{\{P\}}))))\\
\end{eqnarray*}
 \finepatt
\end{patternImpl}
%--------------------------------------------------------

Also in this example, the main parts of this axiom are represented by the concept {\em Transhippment Event} and by the connection between
the container and the vessel events: in this case, this connection is represented by the role {\em hasDischargingVesselEvent}, that allows to pass from the
description of the container itinerary to the one that brought it to the transshipment port.

We have to point out that the instances matching this axiom have to be further elaborated, because it matches all the ships that pass from
the container destination, i.e.,  port $P$ in the example, after the transshipment.
As a simple strategy to prune the suspicious itineraries, one can evaluate the date of arrival of the first vessel to the container destination:
if the date is in the same day, or in very close days, to the one of the container arrival,
%
%\hec{spiegare cosa significa 'close', e' un concetto impreciso}
%
the transshipment was not necessary and the container itinerary can be labeled as anomalous.
% The SPARQL-DL implementation described in Section~\ref{sec:exp}
% extracts the dates of container and vessel arrivals: in this way, it is sufficient to crosscheck them in order to filter false
% positive cases.
% The implementation described in Section~\ref{sec:exp} includes a PL/SQL procedure that, for each container itinerary that satisfies the axiom,
% checks the arrival date of the vessel involved vessel and filter false positive cases.
%
\hec{la procedura PL/SQL tiene conto di tutti e due i casi LOADING and DISCHARGING che vengono discussi nella sezione di experimental evaluation? accennare alle limitazioni sulle date di OWL e DL}
\hec{Potremmo fare una sezione a parte per spiegare come si eliminano i falsi positivi piu' ovvi, partendo dalle considerazioni di Aristide}

%=====================================
\section{Ontology Querying Tools: a Survey}
\label{sec:semtools}
%=====================================
In this section we review the tools and technologies available to query our ontology.
As we discussed above, we can retrieve the trajectories that follow the patterns we are interested in
by checking the DL axioms that formalize such patterns against the ontology, because
axiom checking implicitly creates the classes encompassing the trajectory instances that verify the patterns.
Different DL reasoners can be applied to check the axioms, the most common ones being Pellet~\cite{pellet}, FaCT++~\cite{fact}, Hermit~\cite{hermit} and RacerPro~\cite{racer}.

As an alternative, we can retrieve the trajectory instances by querying the ontology through an ontology Query Language (QL).
This solution augments the expressivity at our disposal for pattern specification, and enables us to test alternative QLs and different QL implementations, possibly
benefiting from improved performance.

%---------------------------------------------------------------------------------------------------------------------
\begin{table*}[ht]
\footnotesize
\begin{center}
\begin{tabular}{|l|l|l|}%p{4cm}|}
\hline
{\bf QL}			&{\bf KBL} 					&{\bf Expressiveness}\\ 			%&{\bf Notes} \\
\hline
\hline
\rowcolor{lightGray}
SPARQL
~\cite{sparql}			&RDF, OWL					&subgraph matching, conjunctive queries\\	%&\\
RQL
~\cite{Karvounarakis_rql}	&RDF					        &subgraph matching\\ 				%&\\
\rowcolor{lightGray}
SeRQL \cite{SeRQL}		&RDF						&subgraph matching\\				%&\\
RDQL  \cite{rdql}		&RDF						&subgraph matching\\				%&\\
\rowcolor{lightGray}
ASK DIG~\cite{dig}		&OWL						&DL atomic queries (TBox/RBox/ABox)\\		%&\\
OWLink protocol
  ~\cite{owllink}		&OWL						&DL atomic queries (TBox/RBox/ABox)\\ 		%&\\
\rowcolor{lightGray}
OWL-QL (DQL)
~\cite{Fikes2004-owlql}		&OWL						&DL atomic queries (TBox/RBox/ABox)\\		%&No implementation\\
%OWL2-QL
%~\cite{ow2ql}
%				&OWL 2.0 DL 					&($DL-Lite_{R}$)				&\\
OWLQ
~\cite{owlq}			&OWL						&DL atomic queries (TBox/RBox/ABox)\\		%&based on Jena framework\\
\rowcolor{lightGray}
SAIQL
~\cite{Kubias07owlsaiql} 	&OWL						&DL atomic queries (TBox/RBox/ABox)\\		%&Part of NEON toolkit\\
nRQL
~\cite{Haarslev04nRQL} 		&OWL						&conjunctive ABox queries\\			%&Integrated with RacerPro\\
\rowcolor{lightGray}
ONTOVQL
~\cite{FadhilH07-ontovql} 	&OWL						&DL atomic queries (TBox/RBox/ABox)\\		%&Visual QL for DL  \\
SQWRL
~\cite{OConnorD08-sqwrl}	&OWL + SWRL					&DL atomic queries + SWRL rules\\		%&support for SWRL rules\\
\rowcolor{lightGray}
SPARQL-DL
~\cite{Sirin07sparqldl}		&OWL						&conjunctive TBox, RBox, ABox queries\\		%&Implemented as extension of Pellet engine\\
SPARQL 1.1
~\cite{sparql11}		&OWL						&conjunctive TBox, RBox, ABox queries\\		%&Entailment regimes\\
\rowcolor{lightGray}
SPARQL-OWL
~\cite{Kollia2011-sparqlowl}	&OWL 	 					&conjunctive TBox, RBox, ABox queries\\ 	%&OWL Direct Semantics entailment. Benchmark for Hermit\\

\hline
\end{tabular}
\end{center}
\caption{\label{tab:ql} Ontology query languages, classified with respect to the language used for the Knowledge Base representation (KBL) and the query language expressivity}
\end{table*}
% %---------------------------------------------------------------------------------------------------------------------
Table~\ref{tab:ql} gives an overview of the existing ontology QLs.
They can be broadly classified into three categories: RDF-based, applying subgraph matching of RDF triples against the ontology graph but lacking DL reasoning capabilities;
DL-based, supporting directly the DL semantics, usually in the form of atomic DL expressions;
and mixed approaches that combine DL expressivity with DL query conjunction.

The most used RDF-QLs is SPARQL, the W3C recommendation for querying triples in RDF graphs through subgraph matching.
DL-based languages enable to express TBox, RBox and ABox queries that can be run directly against OWL files.
For some QL, such as nRQL~\cite{Haarslev04nRQL}, the Racer DL-QL, a limited possibility for query conjunction is also supported.
Other DL-based  approaches augment the QL expressivity providing graphical instruments to specify a query,
like ONTOVQL~\cite{FadhilH07-ontovql},
or integrate the support for rules (i.e., Horn clauses), like SQWRL~\cite{OConnorD08-sqwrl}, which
takes rule antecedents as query specifications.
Finally, the OWLink protocol~\cite{owllink}, which overcomes the ASK DIG interface~\cite{dig} to interact with OWL 2.0 ontologies,
is a reference interface for DL reasoning and querying.

%For what concerns specifically OWL 2, in ~\cite{Perez-Urbina2009-owl2ql} the OWL 2 QL profile, the DL-Lite$_R$ profile of OWL 2, is applied to query rewriting to retrieve data stored in a relational database, while Knorr and Alferes ~\cite{Knorr2011-owl2ql} present an interesting proposal to extend this profile with monotonic rules.

A big step forward towards improving the language expressivity, while preserving decidability and performance, is given by
recent proposals combining the two approaches above,  specifically extending the SPARQL simple entailment based on subgraph matching with with DL reasoning, in particular OWL semantics.
The widest proposal is a recent W3C Candidate Recommendation: SPARQL 1.1~\cite{sparql11}. It
encompasses entailment regimes~\cite{sparql11-entailment} for  RDF, RDFS, RIF Core, D-entailment, OWL Direct and RDF-Based Semantics entailment.
The SPARQL 1.1 specification relies on the work of different communities, including the ones working on SPARQL-OWL~\cite{Kollia2011-sparqlowl}
and SPARQL-DL~\cite{Sirin07sparqldl}.
SPARQL-OWL, in particular, has been implemented extending the engine of the Hermit reasoner (a benchmark is provided, but the source code is not available).
By contrast, a fully functional API for SPARQL-DL~\cite{Sirin07sparqldl} is available. It extends the Pellet~\cite{pellet} query engine, and is currently
a very competitive solution for ontology querying, as we discuss in the experimental evaluation section.
The SPARQL-DL API~\cite{Sirin07sparqldl} and other tools that either support ontology QL or generically enable to query an ontology, are reported in Table~\ref{tab:owltools}.

\begin{table*}[ht]
\footnotesize
\begin{center}
\begin{tabular}{|p{4cm}|p{5.5cm}|p{6.2cm}|}%p{4cm}|}
\hline
{\bf Tool/API} 					&{\bf QL/Expressiveness} 			&{\bf Reasoner}\\
						%&{\bf Notes} \\
\hline\hline
\rowcolor{lightGray}
JENA
  ~\cite{jena} 					&SPARQL			 			&OWL reasoners but only subgraph matching\\
						%&RDF triple abstraction  \\
KAON2
  ~\cite{kaon2}	 				&SPARQL						&Integrated reasoner (OWL Lite, DL safe SWRL, FLOGIC) DIG ASK interface\\
						%&Not OS \\
KAON2 OWL Tools
  ~\cite{owltools}				&SPARQL-DL Lite				 	&OWL-API compliant\\
						%&Wrapper for OWL-API, not maintained\\
\rowcolor{lightGray}
NEON Toolkit
  ~\cite{neon}					&SAIQL						&OWL-API compliant\\
						%&Ontology Editor\\
%ASK DIG	protocol			&DL atomic TBox/RBox/ABox queries		&
						%&\\
Prot\'eg\'e-OWL API
  ~\cite{protegeowl}				&DL atomic TBox/RBox/ABox queries 		&DIG ASK compliant\\
						%&For plugin development  \\
\rowcolor{lightGray}
SQWRL-API
  ~\cite{Kollia2011-sparqlowl}  		&SQWRL	 					&Jess Rule Engine, RacerPro\\% \ec{controllare}\\
						%&Prot\'eg\'e plugin\\
OWL2Query
  ~\cite{owl2query}				&SPARQL-DL$^{NOT}$ 				&OWL-API v. 3 compliant\\
						%&Prot\'eg\'e plugin\\
\rowcolor{lightGray}
RacerPro APIs
  ~\cite{racer}  				&nRQL						&RacerPro\\
						%&jRacer,Racer DIG client \\
OWL-API
  ~\cite{Horridge2011-owlapi}			&DL atomic TBox/RBox/ABox queries		&FaCT++, Hermit, Pellet, CEL {\em (OWL-API v.3 compliant)}, and RacerPro (via OWLLink)\\
						%& \\
\rowcolor{lightGray}
OWLLink API
  ~\cite{owllink}				&DL atomic TBox/RBox/ABox queries 		&RacerPro, OWL-API v.3 compliant reasoners\\
						%&protocol for OWL reasoning systems, evolution of the DIG interface for OWL2.0. Declarative interface for OWL 2.0 reasoner \\
SPARQL-DL API
  ~\cite{Sirin07sparqldl}			&SPARQL-DL  					&OWL-API v.3 compliant reasoners\\

\rowcolor{lightGray}
ORACLE Database Semantic Technologies
~\cite{owlprime}				&RDF,RDFS++,OWLSIF,OWLPrime 			&\\

						%& \\
\hline
\end{tabular}
\end{center}
\caption{\label{tab:owltools} Tools for ontology querying, classified with respect to the query languages (QL) or querying expressivity and the reasoners supported}
\end{table*}
% %---------------------------------------------------------------------------------------------------------------------

Among the tools listed in the table, JENA~\cite{jena} and KAON2~\cite{kaon2} are mainly designed for RDF knowledge bases:
even if they can handle OWL ontologies, reasoning is performed as subgraph matching in JENA, while
KAON2 implements the DIG ASK interface, limited to OWL-Lite for DL reasoning, but partially extended towards SWRL and FLOGIC.
KAON2 OWL TOOLS partially supports SPARQL-DL, but apparently this project is not maintained anymore.

Among the tools specifically designed for OWL ontologies, the OWLink API~\cite{owllink}, the API for the OWLink protocol is
the evolution of the DIG interface for OWL 2.0.
The Prot\'eg\'e-OWL API ~\cite{protegeowl} is an API designed for plugin development, while
and SQWRL-API ~\cite{Kollia2011-sparqlowl} and OWL2Query~\cite{owl2query} are Prot\'eg\'e plugins for ontology querying,
integrating SWRL rules and SPARQL-DL$^{NOT}$, which is SPARQL-DL with
negation as failure, respectively.

NEON~\cite{neon}, RacerPro APIs and SQWRL-API adopt query languages specifically designed for the tools, respectively SAIQL, nRQL and SQWRL.
Of these, the RacerPro API is the most used. However, the supported QL nRQL, as we mentioned above, enables only ABox conjunctive queries; moreover, only the 32bit version of the reasoner
is available and the free license for research has some limitation.

The OWL-API~\cite{Horridge2011-owlapi} is an open source API written in Java that is considered as a reference interface for ontology manipulation. It is widely used and is implemented by several DL reasoners, including FaCT++, Hermit, Pellet, CEL (which are referred to in the table as OWL-API v.3 compliant reasoners), and RacerPro.
It supports directly entailment checking for answering DL atomic queries, but it does not enable to answer conjunctive or SPARQL based queries.

By contrast, as mentioned above, this functionalities are supported by the SPARQL-DL API~\cite{Sirin07sparqldl}, that extends the OWL API to enable conjunctive DL query answering. Moreover, through OWL API, querying can be realized using any OWL API compliant reasoner.

Recently, also mainstream database vendors propose products that combine the ability of databases to handle big amounts of data with the reasoning capabilities
offered by ontology. In its latest version 11g, ORACLE Database includes a module for Semantic Technologies, that supports RDF and OWL files,
with three different vocabularies: RDFS++, which is an extension of RDFS; OWLIFS, OWL with the support of the IF semantics;
and OWLPrime, which is a OWL subset that does not support cardinality property restriction, set operators (union,intersection) and enumeration.
OWLPrime is by far the language that provide the maximum expressivity among those offered by this product,
and OWLPrime expressions can be integrated in SPARQL-like queries that can be specified directly against the database.
Unfortunately, the lack of set expressions does not allow to specify DL axioms with conjunction or disjunction of atomic expression,
limiting the application of this type of products.

%=====================================
\section{Experimental Evaluation}
\label{sec:exp}
%=====================================
%--------------------------------------------------------
\begin{figure*}[t]
\begin{center}
                \includegraphics[width=15cm]{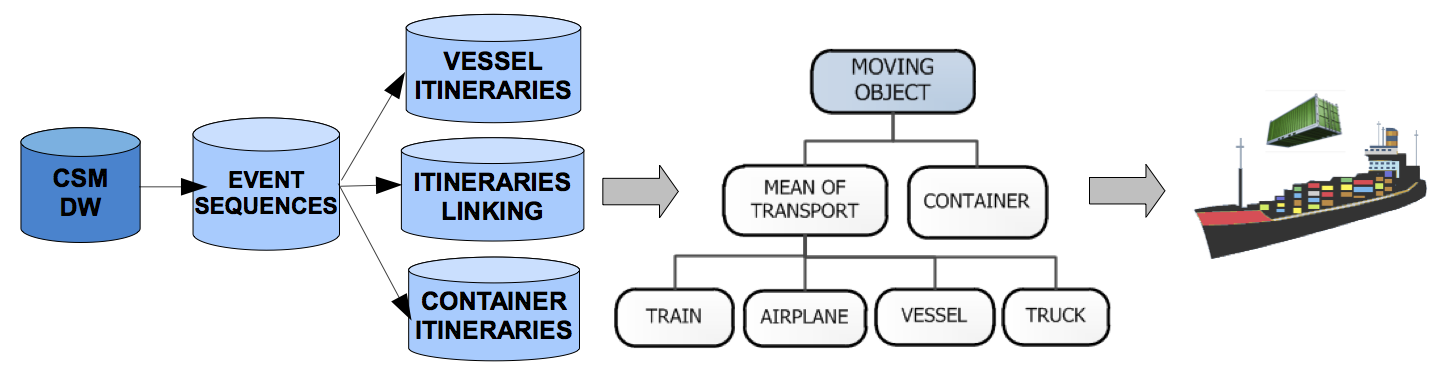}
	\end{center}
\caption{Experimental evaluation process:
(1) Data preparation: data selection, itineraries segmentation, itinerary linking
(2) Ontology population
(3) Ontology querying } \label{fig:evaluation}
\end{figure*}
%--------------------------------------------------------

The experimental evaluation has been organized in three steps, as depicted in Fig.~\ref{fig:evaluation}.
At step (1), we first select the data to process.
We have chosen a sample dataset from the data collected by JRC as part of its container monitoring activity. The dataset
includes 18 millions of {\em Container Status Messages} (CSM). A CSM is a semi-structured text that describes a shipping deed undertaken by carrier companies on a container.
Each CSM includes the position of the container, the operation carried out on it (that we formalize in the MCO as a container event),  its loading status and the vessel used for its transportation.
The initial dataset included CSM referring to 50 thousand containers travelling worldwide for three years, from 2009 to 2012.

% Specifically, Each CSM has the following format (see also Fig.~\ref{fig:seq} for examples of CSM):
% (1) an ISO 6346 container identifier made by: a code for the owner of the container; a code for the
% category of container; a unique serial number; and a check digit \footnote{Given for instance container identifier ABCD1234567, ABC identifies the carrier company, D is the container category; 123456 is a serial number and 7 is a check digit.}.
% (2) the text description of the event occurring to the container, such as loaded, discharged, etc.;
% %\ec{check event codes defined in the ANSI X.12 or UN EDIFACT standards};
% (3) the date when the event occurred;
% (4) the place, usually a port, where the event took place;
% (5) the loading status of the container (empty or full);
% (6) depending on the event type, a vessel identifier.
%
% Extract-Transform-Load (ETL) procedures are applied to CSMs in order to  prune errors and to harmonize them with respect to standard taxonomies for for events, loading status, locations and vessels and then stored in a database.

During the pre-processing phase in step (1), we segment CSM sequences to extract container itineraries, identifying container shipments and vessel
trips.
As a result of the segmentation phase, more than 290 thousand container itineraries and more than 43 thousands %43698
vessel trips have been identified.
%We need to link each container trip with the corresponding vessel trips.
Since usually more than one vessel is used for accomplishing a container shipment, and every vessel transports
in a single trips thousands of containers, we need to map every part of a container itinerary with the corresponding vessel trips. This concludes the
pre-processing phase.

We populate the MCO at step (2) with the itineraries and the related information. The MCO has been implemented in
OWL-DL, the description logic sublanguage of the Web Ontology Language OWL~\cite{owl}, according to the design described in Section~\ref{sec:mco}.
OWL is widely used for ontology definition, therefore a lot of tools and libraries are available for ontology editing, population and
visualization and querying (cf. Section~\ref{sec:semtools}). Moreover, it includes semantic features to enhance reasoning, in particular ontology axioms, that we use to express suspicious itinerary patterns.
Among the available tools, we chose the Jena Java API~\cite{jena} for populating the ontology.

To have a more meaningful evaluation of the approach, in particular in terms of performance scalability, we run different tests using four ontologies of different sizes,
randomly created starting from the initial dataset, that contain, respectively: 100589, 153816, 207356 and 260637 individuals.
%that contain, respectively: 5000, 10000, 15000 and 20000 container itineraries \ec{cambiare dando il numero totale degli individui dell'ontologia}.
To have an insight on the complexity of the ontology, we can consider the number of other types of individuals, namely container and vessel itineraries, containers, vessels, and ports, as summarized in Table~\ref{tab:dataset}. Notice for instance that, while the number of containers increases more or less proportionally to the number of container itineraries, which is our reference dimension for the experimental evaluation, the increase in the number of vessels remains limited.
This phenomenon is more evident when considering the ports traversed by the itineraries. The fact that the number of ports remains bounded is an advantage for our application, because in the evaluation of the axioms, we have to scan iteratively all the ports the containers passed through, therefore the number of ports in the dataset can become very easily a bottleneck for the application.
By contrast, if the axiom has acceptable performance with a limited number of container itineraries,
we can expect reasonable processing time even with a bigger number of shipments, because the number of ports does not increase proportionally.
%\pv{non mi convince, perche' questo avviene solo in questa sperimentazione, non e' detto che avvenga in generale}

We expect this consideration applies as well in other application domains, for example the locations crossed by itineraries do not increase proportionally when considering a bigger dataset of trajectories.
%---------------------------------------------------------------------------------------------------------------------
\begin{table*}[ht]
\scriptsize
\begin{center}
\begin{tabular}{|c|c|c|c|c|c|c|}
\hline
Ontology 	& OWL individuals 	& Container itineraries 	& Containers  	& Vessels 	& Ports   \\
\hline
owl5K  		& 100589		& 5000 				& 4763  			& 841 		& 565     \\
owl10K 		& 153816 		& 10000 			& 9203 					& 960 		& 593     \\
owl15K 		& 207356 		& 15000 			& 13264 				& 1023 		& 604     \\
owl20K 		& 260637 		& 20000 			& 17012 				& 1078 		& 618     \\
\hline
\end{tabular}
\end{center}
\caption{\label{tab:dataset} Number of ontology individuals in the different OWL files used for the experimental evaluation
for concepts Container itinerary, Containers, Vessels and Ports.}
\end{table*}
% %---------------------------------------------------------------------------------------------------------------------

At step (3), we query the MCO against a set of DL-queries that implement the anomalous itinerary axioms we formalized in Section~\ref{sec:axioms}.
We tested different ontology APIs, languages and reasoners: the OWL-API \cite{Horridge2011-owlapi} and SPARQL-DL \cite{Sirin07sparqldl} DL-query languages, combined with
Pellet~\cite{pellet}, Hermit~\cite{hermit}, FaCT++\cite{fact}.

%==============================================
\subsection{Data selection and pre-processing}
\label{sec:datapreparation}
%==============================================

%-------------------------------------------------------------------------------------------------------------------------------------------------------------%
\begin{table*}[ht]
\centering
\scriptsize
\begin{tabular}{|c|c|c|c|c|c|c|}
\hline
\bfseries CSM identifier &\bfseries Container identifier & \bfseries Time & \bfseries Event& \bfseries Location& \bfseries Loading status& \bfseries Vessel\\
\hline\hline
12345	&ABCD1234567 &27 May 2010	&Received at Origin		&Shangai (CN) 		&Empty	&--\\
12346	&ABCD1234567 &27 May 2010	&Gate In			&Shangai (CN) 		&Full	&--\\
12350	&ABCD1234567 &30 May 2010	&Loaded/Ramped			&Shangai (CN) 		&Full	&Aurora\\
12365	&ABCD1234567 &15 Jun 2010	&Discharged/Deramped		&Port Kelang (MY) 	&Full	&--\\
12366	&ABCD1234567 &17 Jun 2010	&Loaded/Ramped			&Port Kelang (MY) 	&Full	&Dawn\\
12381	&ABCD1234567 &03 Jul 2010	&Discharged/Deramped		&Antwerpen (BE) 	&Full	&--\\
12399	&ABCD1234567 &09 Jul 2010	&Gate Out			&Antwerpen (BE) 	&Full	&--\\
12455	&ABCD1234567 &16 Jul 2010	&Final Destination		&Antwerpen (BE) 	&Full	&--\\
12484	&ABCD1234567 &20 Aug 2010	&Received at Origin		&Antwerpen (BE) 	&Empty	&--\\
12545	&ABCD1234567 &23 Aug 2010	&Gate In			&Antwerpen (BE) 	&Full	&--\\
12555	&ABCD1234567 &24 Aug 2010	&Loaded/Ramped			&Antwerpen (BE) 	&Full	&Sun\\
\hline
\end{tabular}
\normalsize
\caption{Example of container sequence for container ABCU1234567
 \label{tab:seq}}
\end{table*}
%-------------------------------------------------------------------------------------------------------------------------------------------------------------%

For each container in the CSM dataset, we extracted the corresponding {\em event sequence}, %i.e., the ordered sequence of CSM it has been involved in.
%Each sequence
which details the shipment history of a single container.
%For each container, only one sequence of container events exists in the dataset.
An example of container sequence is reported in Table~\ref{tab:seq}.
Each line in the table represents a CSM, which is composed by: a CSM identifier; an ISO 6346 container identifier\footnote{In ISO 6346 identifier ABCU1234567, ABC identifies a carrier company, D is a container category; 123456 is a serial identification number and 7 is a check digit.}; the date when the event occurred;
textual description; the place, usually a port, where it took place;
the loading status of the container (empty or full); depending on the event type, a vessel identifier.

% The schema of the container sequence dataset has the following relational schema\footnote{The types used in the definition of the
% dataset  are  ORACLE\textregistered~\cite{oracle} datatypes.}:
% \begin{verbatim}
% MOVEMENT		NUMBER(38)
% BIC_CODE		NUMBER(38)
% CONTAINER_ID		NUMBER(38)
% TIME			DATE
% LOCATION		NUMBER(38)
% EVENT			NUMBER(38)
% LOADSTATUS		NUMBER(38)
% VESSEL		NUMBER(38)
% \end{verbatim}
% where: {\tt MOVEMENT} is a container event identifier;
% {\tt TIME}  is a date of event; and {\tt CONTAINER\_ID},
% {\tt LOCATION},  {\tt VESSEL}  are container, location and vessel identifier, respectively; {\tt BIC\_CODE} identifies the container owner
% according to ISO 6346 standard, as assigned by the International Container Bureau \cite{bureau};  {\tt EVENT} and  {\tt LOADSTATUS} identify
% the container event type and the loading status.

Each container sequence is then processed to extract the container and vessel itineraries, as described next.

%---------------------------------------------------
\subsubsection{Reconstructing Container Itineraries}
%---------------------------------------------------

%--------------------------------------------------------
\begin{figure}[t]
\begin{center}
                \includegraphics[width=8cm]{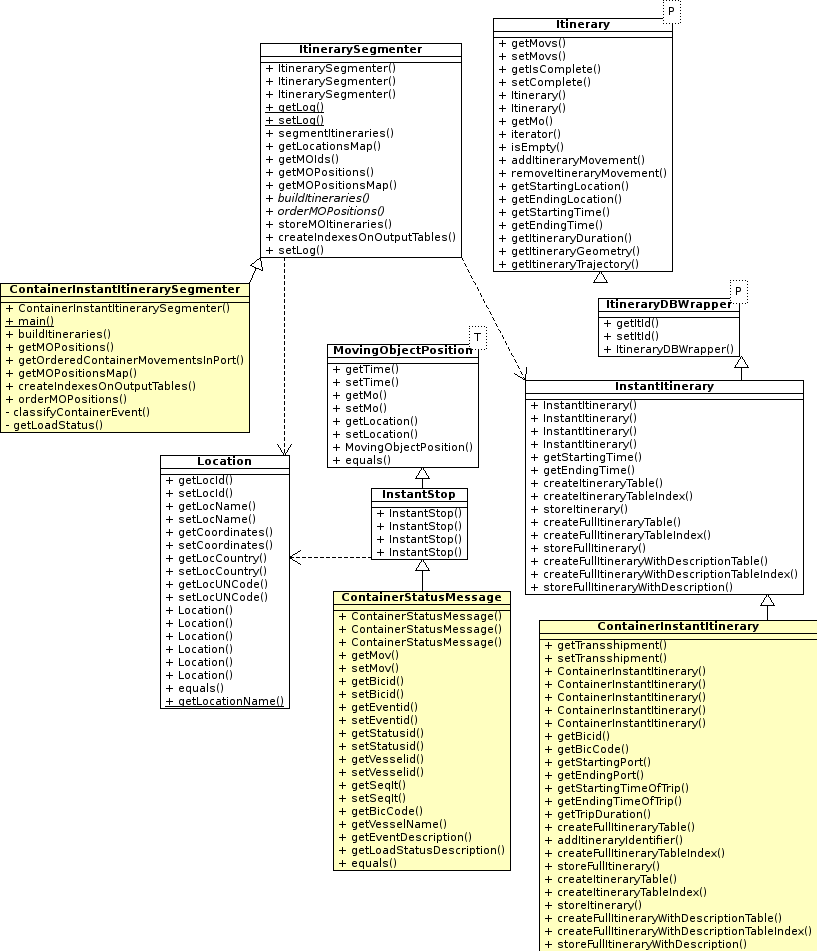}
	\end{center}
\caption{UML Class diagram of the API for itinerary segmentation} \label{fig:acid}
\end{figure}
%--------------------------------------------------------

The itinerary segmentation is implemented in Java, and leverages the semantics of container events,
as defined in the ontology excerpt reported in Fig.~\ref{fig:events}. The class diagram of the API is reported in Fig.~\ref{fig:acid}.
Specifically, we segment every container event sequence among different shipments.

Ideally, an itinerary is composed by the following phases, corresponding to the five main categories of events described in Section~\ref{sec:container}:
\begin{enumerate}[-]
 \item  Begin of Trip;
\item Container Export;
\item an optional sequence of Container Transshipments;
\item Container Import;
\item End of Trip.
\end{enumerate}
% (1) Begin of Trip;
% (2) Container Export;
% (3) an optional sequence of Container Transshipments;
% (4) Container Import;
% (5) End of Trip.
Given for instance the sequence in Table~\ref{tab:seq}, it includes two itineraries for container ABCD1234567: the first starting at Shangai
in China on the 27th of May and ending at Antwerpen in Belgium on the 16th of July; and the second, which is partial, starting at Antwerpen the 20th of August.
Note that we can have gaps in the event sequence, therefore the segmentation algorithm can produce partial itineraries, or merge different itineraries in a single one.
To partially overcome this issue, the algorithm takes into account also events %which are included in the standard classification,
that do not describe a container movement but are deeds occurring to prepare the container for the shipping at the source port or to complete it at the port of destination
(e.g.,  released to shipper for cargo stuffing, empty returned).
These events, complemented with the loading status of the container, help put a container in a specific port at the begin and at the end of a shipment,
and define more precisely the temporal period a container spends in a port. %; therefore, they are included in the container itinerary, to provide a more complete as possible description of the shipment.

\subsubsection{Reconstructing Vessel Trips}
%---------------------------------------------------
Vessel itineraries are extracted from the same dataset of container sequences we processed above.
Indeed, vessel routes are implicitly defined by CSMs that can include also the names of the vessels used for the container transportation.
Typically, a vessel transports many containers in a single trip between two ports, hence its movements can be inferred by considering
CSM of different containers. In this case, we are likely to overcome the issue of incomplete container sequences.

For each vessel in the dataset, we aggregate container events with respect to their occurrence in each port at a specific time,
obtaining the temporal interval during which the vessel stopped in each port.
Ordering such interval-based vessel events, we obtain a sequence of events for the vessel, with the events dates and locations,
from which we infer the event description, i.e., departure or arrival. Vessel itineraries are extracted from vessel sequence, considering them made by pairs of {\em departure} and {\em arrival} vessel events.

%---------------------------------------------------
\subsubsection{Binding Itineraries to Trips}
%---------------------------------------------------

Once container and vessel itineraries have been reconstructed, we proceed to link them relying on transshipment events.
Transshipments play a fundamental role in both the anomalous axioms described in Section~\ref{sec:axioms}, therefore, in order to detect the
corresponding anomalous patterns, we need to set correctly the roles involved in the transhippment specification. These ones are not
explicit in the dataset, but should be set explicitly in the ontology.
Therefore, we connect every discharging container event
with the arrival event of the corresponding vessel that occurs immediately {\em before} its discharge; similarly, every loading container event
with the vessel departure that happens immediately {\em after} its loading. The results of this procedure, which has been implemented in ORACLE PL/SQL, are stored back in the database.

\subsection{Ontology population} \label{subsec:Ontology population}
%=====================================
%--------------------------------------------------------
\begin{figure}[t]
\begin{center}
\includegraphics[width=8.5cm]{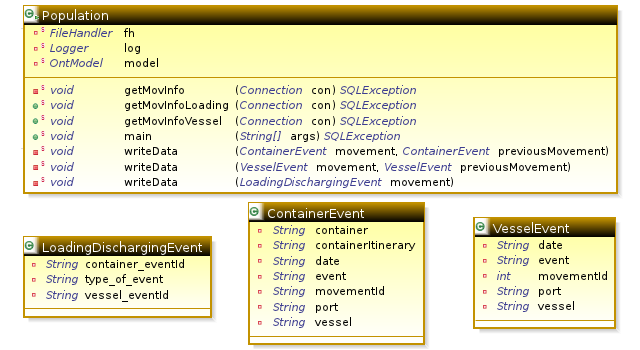}
\end{center}
\caption{Classes for Knowledge Base Population}\label{fig:population}
\end{figure}
%--------------------------------------------------------
We use the Jena \cite{jena} framework to obtain four populated ontology files, described in Table~\ref{tab:dataset},
that has to be queried to detect anomalous itineraries.
To reach this goal, we implemented an ad-hoc Java package,
whose design is illustrated in Fig.~\ref{fig:population},
and whose main classes
describe the domain knowledge base for the MCO.
For sake of simplicity, we show in Fig.~\ref{fig:population} only the attributes of these classes.
%: specifically, the containers and vessels events are represented by Java classes.
The population of the ontology is fulfilled by the class {\tt Population.java} (see~Fig.\ref{fig:population})
which, relying on Jena, builds the corresponding objects to insert them into the the ontology source file.
\subsection{Detecting anomalous itineraries} \label{subsec:Detecting anomalous itineraires}
%=====================================

% The goal of this last phase is to understand which combination of reasoner and query language is more suitable to query an ontology
% against the axioms in Section \ref{sec:axioms}.
%The goal of this phase is the detection of the anomalous itineraries.

% For limitations with the license, we have not used the reasoner RacerPro~\cite{racer}, and we have not taken into
% account KAON2~\cite{kaon2} because
% it does not support nominals.

We have started to test the axioms by considering the Java OWL-API~\cite{Horridge2011-owlapi} interface. Among the compatible reasoners,
we have tested FaCT++, HermiT, and Pellet. RacerPro has platform limitations and its free license for research has limitations.
After having collected their performance in terms of time used to get the positive cases, we have searched for other tools in order to get better
results. We have taken into account the SPARQL-DL\cite{Sirin07sparqldl} engine in Pellet, and the SPARQL-DL implementation by Derivo.

All the tests have been done using a PC with a 64
bits processor: Intel(R) Xeon(R) CPU E5620, equipped with 133MHz of clock, reserving
 5 Gb of RAM to the process.

In the following, for each suspicious pattern we show the steps we have followed in our experimentation, and the performance of our tests.

%=====================================
\subsubsection{ Detecting Unnecessary Transshipments} \label{subsec:querying unnecesstrans}
%=====================================

%----------------------------------------------------------------------------------------
\begin{figure}[!htb]
\begin{center}
\begin{tikzpicture}
\begin{tiny}
\begin{semilogyaxis}
[ymin=0,ymax=50000,
height=5cm,
width=8cm,
ymajorgrids=true,
legend style={legend pos=north west},
% xtick pos=left,
%title={\small \label{fig:unnecessGraph} \trans performances},
title={},
xlabel={{number of container itineraries}},
ylabel={minutes (ln scale)},
symbolic x coords={5000,10000,15000,20000},
xtick=data,nodes near coords,
nodes near coords align=%
{vertical}]

\addplot[thick,mark=*,color=black]
%pellet owl-api
coordinates
{(5000,180.33)
(10000,647.20)
(15000,1600.5)
(20000,3500.9)};
%fact ++ owlapi
\addplot[thick, mark=square*,mark options={fill=blue},color=blue]
coordinates
{(5000,11.5)
(10000,44.5)
(15000,89.5)
(20000,150.5)
};
\addplot[thick,mark=triangle*, color=red]
coordinates
% sparql-dl
{(5000,3.2)
(10000,4.25)
(15000,5.5)
(20000,7.9)};

\legend{\tiny Pellet \& OWL-API, \tiny FaCT++ \& OWL-API, \tiny Pellet \& SPARQL-DL}

\end{semilogyaxis}
\end{tiny}
\end{tikzpicture}
\caption{Performance of \trans~detection}\label{fig:unnPerf}
\end{center}
\end{figure}
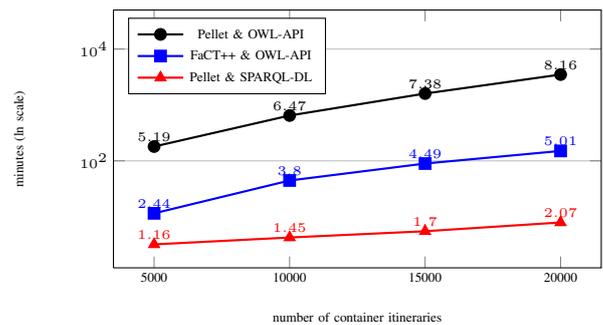
%----------------------------------------------------------------------------------------

%----------------------------------------------------------------------------------------
\begin{table}[t]
\scriptsize
\setlength{\tabcolsep}{3pt}
\centering
\begin{tabular}{|l|r|r|r|r|}
\hline
  & \multicolumn{4}{c|}{Computation time (minutes)} \\
\hline
Reasoner/Interface & \emph{owl5K} & \emph{owl10K} & \emph{owl15K} & \emph{owl20K}   \\
\hline
% Pellet/OWL-API & 180m22.592s & 647m12.386s & 1600m55.362s & 3005m56.005s\\
% Pellet/SPARQL-DL & 3m19.15s & 4m15.89s & 5m36.006s & 6m57.762s\\
% Fact++/OWL-API & 11m29.770s & 44m7.557s & 89m29.171s & 150m27.176s\\
Pellet \& OWL-API & 180 & 647 & 1600 & 3005\\
FaCT++ \& OWL-API & 11 & 44 & 89 & 150\\
Pellet \& SPARQL-DL & 3 & 4 & 5 & 6\\
\hline
\end{tabular}
\caption{Performance of \trans~\protect\newline
detection for different reasoners and interfaces}
\label{tab:unnPerf}
\end{table}
%----------------------------------------------------------------------------------------

In Table~\ref{tab:unnPerf} and in Fig.~\ref{fig:unnPerf}, we show the most performant results of the experimental phase with the \trans~axiom. We have started
our experimentation relying on OWL-API, and we have exploited it by developing a Java package called {\tt itin\-er\-aries.\-que\-ry},
that is based on Matthew Horridge's example code in~\cite{Horridge2011-owlapi}.

In the core class of this package, we extract from the database all the ports where the containers passed through,
and we test the axiom of Section~\ref{sec:axioms} against every port.
To reach this goal, the axiom has been rewritten into Manchester syntax for OWL~\cite{HorridgeP08}. We have tested three reasoners:
HermiT, Pellet and FaCT++. We have found that FaCT++ is by far the reasoner that performs better with OWL-API.
On the other hand, we stopped testing Hermit after having realized that, in the case with the smallest dataset,
its computation took more than twice as long as the one with Pellet.

However, the main problems with the OWL-API pure approach are the slowness of the computation, and the necessity of another mechanism in order to
clean the itineraries found by selecting those that have compatible arrival dates (see Section~\ref{sec:axioms} for details).
Actually, OWL-API does not allow us to extract this information.

As an alternative, we considered SPARQL-DL~\cite{Sirin07sparqldl}: it is an expressive
language for querying OWL-DL ontologies, and allows us to extract the dates that are necessary to get the
real suspicious itineraries. Moreover, Pellet is equipped with an engine that can speed up the performance with this tool.
In Table~\ref{tab:unnPerf} and in Fig.~\ref{fig:unnPerf}, we can see that the results with Pellet and SPARQL-DL
are better performing than the pure OWL-API approach.
After this test, we have decided to test a generic implementation of SPARQL-DL, and we have considered the one by Derivo.
Since their SPARQL-DL query engine is settled on top of the OWL-API, we have tried to combine its use with the more performant reasoner
with OWL-API, that is FaCT++ according to our tests. Unfortunately, in this test the performance has been very bad and we have stopped it
when we have realized that it would not have terminated the execution in a reasonable time. The code of the experiment can be seen in
the appendix.

By looking at the experiments results, we can see that the combination of SPARQL-DL and Pellet is by far the best one in terms of time: for example,
if we consider the case of 10000 itineraries, we have an improvement of more than 99\% of time with respect to the OWL-API and Pellet approach. If
we take into account the other cases, we have improvements of the same order of size.
%It is obvious that we can not give a general opinion on tools performance by basing our considerations only on these tests, but we have to remark that the combination of SPARQL-DL and Pellet enables us to detect the anomalies in a reasonable time.
Moreover, we have to observe that SPARQL-DL enables to compare dates, hence its use eliminates
the need of a post-processing phase for eliminating false positive cases. Hence, it seems the most appropriate to analyse itineraries of this kind.
%The next experiment5gives a similar result.

%=====================================
\subsubsection{Detecting Loops} \label{subsec:querying cycle}
%=====================================

%--------------------------------------------------------------------------------------------------------------------------------------------------------------------------
\begin{center}
\begin{figure}
\tiny
\begin{tikzpicture}
\begin{semilogyaxis}
[ymin=0,ymax=10000000,
height=6cm,
width=8cm,
ymajorgrids=true,
legend style={legend pos=north west},
% xtick pos=left,
title={},
xlabel={\tiny{number of container itineraries}},
ylabel={\tiny minutes (ln scale)},
symbolic x coords={5000,10000,15000,20000},
xtick=data,nodes near coords,
nodes near coords align=%
{vertical}],

%pellet owlapi
\addplot[thick,mark=*,color=black]
coordinates
{(5000,1444.5)
(10000,2555.5)
(15000,3951.9)
(20000,5315.3)
};

\addplot[thick,mark=triangle*, color=red]
[draw=red]
coordinates
%pellet and sparql-dl 1
{(5000,39.9)
(10000,512.2)
(15000,588.5)
(20000,658.5)
};

\addplot[thick,mark=diamond*,color=violet,nodes near coords align=%
{below}
]
[draw=violet]
coordinates
%pellet and sparql-dl 2
{(5000,3.5)
(10000,5.5)
(15000,8.7)
(20000,10.3)};

\addplot[thick,mark=star,color=orange]
[draw=purple]
%pellet owlapi
% 5000:   5m49.588s
% 10000: 7m19.239s
% 15000:  9m8.555s
% 20000:	40m46.455s
coordinates
{(5000,5.49)
(10000,7.19)
(15000,9.8)
(20000,40.46)
};
\legend{\tiny Pellet\&OWL-API, \tiny Pellet\&SPARQL-DL(1), \tiny Pellet\&SPARQL-DL(2),  \tiny Pellet\&SPARQL-DL(3)}
\end{semilogyaxis}
\end{tikzpicture}
\caption{Performance of \cycle~ detection.}\label{fig:Loop2Graph}
\end{figure}
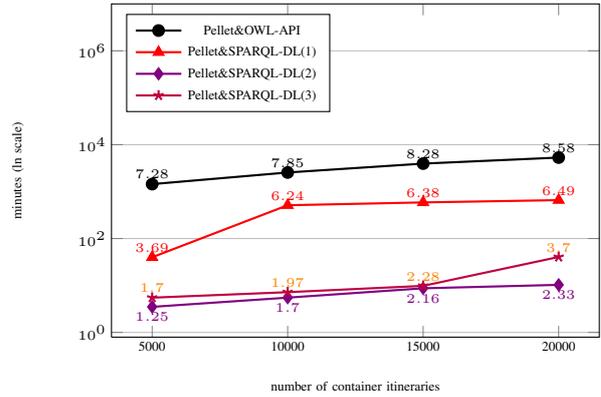
\end{center}
%\captionof{figure}{Results with the \cycle~pattern and the new formalisation}
%---------------------------------------------------------------------------------------------------------------------------------------------------

%---------------------------------------------------------------------------------------------------------------------------------------------------
\begin{table}[ht]
\scriptsize
\setlength{\tabcolsep}{3pt}
\centering
\begin{tabular}{|lc|r|r|r|r|}
\hline
  & \multicolumn{5}{c|}{Computation time (minutes)} \\
\hline
\multicolumn{2}{|l|}{Reasoner\&API} & \emph{owl5K} & \emph{owl10K} & \emph{owl15K} & \emph{owl20K}   \\
\hline
% Pellet/OWL-API & 1444m37 & 2555m25 & 3951m94 & 5315m21.602s\\
% Pellet/SPARQL-DL(1) & 39m44.205s & 512m7.795s & 588m5.321s & 658m30.700\\
% Pellet/SPARQL-DL(2) & 3m37.167s & 5m36.872s & 8m15.272s & 10m23.417s \\
\multicolumn{2}{|l|}{Pellet \& OWL-API} & 1444 & 2555 & 3951 & 5315\\
\hline
\multirow{3}{*}{Pellet \& SPARQL-DL} & [1] & 39 & 512 & 588 & 658\\
& [2] & 3 & 5 & 8 & 10 \\
& [3] & 5 & 7 & 10 & 40 \\
\hline
\end{tabular}
\caption{Performance of Loop detection.\protect\newline
For the rows {\em Pellet \& SPARQL-DL}, we have cases:\protect\newline
 [1] without date filter;\protect\newline
 [2] with date filter;\protect\newline
 [3] with date filter and intermediate ports.}
\label{tab:loopPerf}
\end{table}

%---------------------------------------------------------------------------------------------------------------------------------------------------

%
In Table~\ref{tab:loopPerf} and in and in Fig.~\ref{fig:Loop2Graph}, we have the fastest performance of the experimental phase with the \cycle~axiom.
We obtained an acceptable performance with OWL-API only when combined with Pellet: actually, in the other cases (involving HermiT and FaCT++)
we were obliged to stop the tests because of the slowness of their computation.

We have also tested the SPARQL-DL version of the axiom, ({\em Pellet \& SPARQL-DL(1)} in the Figure),
and we have found an improvement in performance.

However, exploiting the ability offered by SPARQL-DL
to compare dates, we were able to test another formalisation of the query: in this version, it considers containers
loaded on a ship that goes back to its port of start and, after this fact, it is
discharged. This formalization improves by far the performance ({\em Pellet \& SPARQL-DL(2)} in  Fig.~\ref{fig:Loop2Graph}) to be compared with the
previous versions.

By exploiting the same ability, we have implemented also the other version of the query, that matches the itineraries when a container goes back to an
intermediate port before reaching its final destination.  The performance of this experiment is labelled by {\em Pellet \& SPARQL-DL(3)}
in Table~\ref{tab:loopPerf} and in  Fig.~\ref{fig:Loop2Graph}. The code of the experiment can be seen in the appendix.

By looking at the experiments results, we can see that also in this case
that the joint use of SPARQL-DL and Pellet is by far the best one in terms of time: for example,
if we consider the case of 10000 itineraries, the performance of Pellet \& SPARQL-DL(1)
improves of almost 80\% the time with respect to the OWL-API and Pellet approach. Moreover, the possibility to compare dates enables us
to rewrite the query in a different way: this can cause a further improvement of the performances, and this is what happens with the Pellet \& SPARQL-DL(2)
query version. This improvement of the performance has given us the reason to implement and test the  Pellet \& SPARQL-DL(3)
query version. From these tests, we can deduce that the combination of SPARQL-DL and Pellet seems one of the most indicates
to implement our methodology.

We remark that, while in the \trans~experiment Fact++ seemed to be the most promising
reasoner to be used with the OWL-API, in this case Pellet has obtained better performance. Relying only on
the OWL-API, it would be very difficult to choose the best reasoner for the application. However, the solution that combines Pellet and SPARQL-DL is efficient in both cases.

% If
% we take into account the other cases, we have improvements of the same order of size. It is obvious that we can not give a general opinion on tools
% performance by basing our considerations only on these tests, but we have to remark that the combination of SPARQL-DL and Pellet enables
% us to detect the anomalies in a reasonable time. Morevoer, we have to observe that this tool can compare dates, hence its use eliminates
% the need of a post-processing phase. Hence, this tool seems the most appropriate to analyse trajectories of this kind. The next experiment
% gives a similar result.

%=====================================
\section{Discussion and Conclusions} \label{sec:conclusion}
%=====================================

% In the previous sections we have shown a description logic formalism that supports the discovery of suspicious patterns from maritime container
% trajectories stored as MKB instances.
%
% With respect to other case studies addressed in the literature \cite{BaglioniMRTW09}, the domain knowledge encoded by MKB and the reasoning
% necessary to discover anomaly are very complex. In particular, the detection of suspicious patterns requires multiple itineraries be analysed
% at the same time, i.e., container itineraries and vessel routes. Indeed, in \cycle~axiom the itinerary of one container is evaluated against
% one vessel itinerary, while in \trans~axiom the same itinerary is compared with two vessel routes. To partially reduce the inherent complexity
% of the domain, in MKB we introduced redundant relationships between concepts to avoid the specification of long paths in axioms. This simplifies
% their evaluation, but particular care is necessary when populating the corresponding roles in MKB. To avoid inconsistencies, this step can be
% formalised by triggering a set of rules written in Semantic Web Rule Language (SWRL, \cite{swrl}), that extends  OWL axioms  to support Horn-like rules.

In this paper, we have shown a semantic approach for pattern discovery in trajectories that, relying on ontologies, enhances moving object information with event semantics. Our methodology includes a top-level ontology for modelling moving object trajectories, that can be extended to formalize the semantics of a specific application domain. The domain ontology can be queried to search for trajectories following given patterns. These can be formalized as ontology axioms, or specified as DL queries using some ontology query languages. We have validated our approach in a real world scenario, evaluating different implementation solutions.

The main asset of this approach is the possibility to define concepts and properties by exploiting the ontology expressivity and its capability of abstracting the entities of the application domain. In particular, axioms formalizing patterns may be expressed in terms of high-level semantic concepts, abstracting from the specific modelling adopted to represent the domain.
This is a remarkable feature in heterogeneous domains like the one we considered for testing, because it enabled us to refer to the standard events classes defined in the ontology instead of referring the specific events defined by carrier companies using their own vocabulary.

Moreover, this approach enables to use a DL reasoner for building an automatic system for the characterization of different itineraries in terms of the user's needs. The approach is robust because the decidability of axiom evaluation is guaranteed by the robusteness of the DL formalism.

%Another drawback is their limited expressive power, that sometime forces the development of artificial domain formalisations.
% For example, it is not possible to refer to variables in the axiom specification, therefore in  Section~\ref{sec:axioms} we had to
%  specify each shipping port as a full semantic concept.
It is worth mentioning that, for application domains requiring more complex formalization, we can further improve the representation language expressivity
%However, weakening the decidability constraint, we can overcome this difficulty
using formalisations such as OWL and SWRL\cite{swrl},  to enable the use of variables and express equality comparison between instances.
However, this entails weakening the decidability constraint.

% This formalisation can be complemented by other evaluations that can be applied as well, such as
% the recording and the comparison of mean time to accomplish certain handling time in ports.

% The ontology-driven enrichment of moving object itineraries is a promising approach in discovering anomalous itinerary patterns. In particular,
% the formalisation we have proposed may be extended to search for other types of anomalous patterns: the way to do it is to define
% the corresponding axioms,  then to use a reasoner such as Pellet~\cite{pellet} to evaluate them.

% This approach might saw the seeds of a methodology to solve similar problems for moving object itineraries
% in other application domains, especially in GISScience field:

% The exploitation of the semantic hidden in trajectories of moving objects does not involve only Maritime Surveillance
% area. Indeed, the discovery of patterns in itineraries is an underlying problem in GISScience, to retrieve traffic patterns and way-finding for urban
% modelling, etc.
% Moreover, this work offers a formal study to develop an extended formalisation of search patterns, and it may be applied to discover anomalies in
% spatio-temporal sequences formalised as itineraries, for example intrusion detection in Secure areas.

The use of an ontology to describe the behaviour of movement has also some drawbacks. In particular, scalability with a large datasets is an open issue.
In the case of maritime surveillance and security, the search of suspicious patterns may involve the analysis of several thousands of records,
therefore we have to take into consideration scalability when chosing the approach to apply.

As we have discussed in Section~\ref{sec:semtools}, recently,
%A potential solution to this problem
%(that is inherent to the DL approach and well known in the Formal Logic research field)
%is the use of
reasoning engines %which are
specifically designed to handle big knowledge bases have been presented\cite{oracleSemantics}.
%formalised using standard semantic languages such as OWL and Resource Description Framework (RDF) and RDF Schema (RDFS) \cite{rdfs,rdf}.
%For example, Oracle Database Semantic Technologies \cite{oracleSemantics} provides different tools (RDFS++, OWLSIF and OWLPrime) to
%infer new knowledge from semantic enabled repositories.
However, even if this products are a potential solution to the scalability issue, currently these technologies are not mature enough, because their expressivity is very limited, and lack of fundamental DL operations (e.g., OWLPrime does not provide union and intersection \cite{oracleSemantics}, which are necessary for axiom evaluation).
%However, such products currently lack of fundamental DL operations, such as union and intersection \cite{oracleSemantics}, which are necessary for axiom evaluation.

Another way to face up with the scalability issue might be the development of pre-processing procedures to reduce the size of the
dataset, providing the DL reasoner with a smaller knowledge base input. The same approach has been adopted in \cite{Bogorny2010},
where an input dataset of touristic trajectories is first pre-processed with a set of data mining procedures to discover a bunch
of data-mining patterns; only after this step, such patterns are loaded in the knowledge base to reason on them.

However, in the test scenario we have considered, we showed that the combined use of Pellet and SPARQL-DL API is efficient even when considering datasets with thousands of itineraries and instances, and we can obtain even better performance when applying some a priori filtering directly in the DL query specification.

We remark that at the moment our approach is related only to complete itineraries:
a possible extension of this work will be to integrate data mining technologies for managing incomplete itineraries. Moreover, since a peculiarity
of such technologies is to discover implicit semantics, we can rely on them to manage unexpected patterns.

As for the future work, we plan to investigate the employment of OWL and SWRL formalism in order to increase the expressiveness of our approach.
Moreover, we plan to study the development of pre-processing procedures to reduce the size of the initial dataset. We are currently developing
a pre-processing module to address container itineraries that includes also non explicit events, such as the container passing in a port without being handled.
These can be retrieved by reasoning vessel events, defined relying on other containers that travel on the same vessel.

%=====================================
% BIBLIOGRAPHY
%=====================================
\scriptsize
\bibliographystyle{spbasic}
\bibliography{semTrj}
%=====================================

%=====================================
\section*{APPENDIX A} \label{sec:app}
%=====================================

\subsubsection*{Querying \trans}
%Here we give the corresponding formalisation of pattern \trans~with OWL-API and Clark \& Parsia SPARQL-DL syntaxes:
\begin{dlquery}{\bf \trans in OWL-API}\\
\begin{scriptsize}
\lf{Maritime\_Container\_Itinerary}~\lf{and}\\
\lf{hasCDestionationPort}~\lf{value}~\lf{P}~\lf{and}\\
\lf{hasContainerEvent}~\lf{some}~\lf{(Transhipment\_Event}~\lf{and}\\
\lf{hasDischargingVesselEvent}~\lf{some}~\lf{(hasNextVesselEvent}~\lf{some}\\
\lf{(Event}~\lf{and}~\lf{hasVPort}~\lf{value}~\lf{P)))}
\end{scriptsize}
 \finepatt
\end{dlquery}

\begin{dlquery}{\bf \trans in SPARQL-DL}\\
\begin{scriptsize}
\lf{SELECT~DISTINCT~?c~?endCI~?vesStop~WHERE~\{~}\\
\lf{?c~a~st:Container\_itinerary~.~}\\
\lf{?c~st:hasEndTime~?cd~.~}\\
\lf{?c~st:hasCIDestinationPort~st:port~.~}\\
\lf{?c~st:hasContainerEvent~?t~.~}\\
\lf{?t~rdf:type~?eventClass~.~}\\
\lf{?eventClass~rdfs:subClassOf~st:Transshipment\_Event~.~}\\
\lf{?t~st:hasDischargingVesselEvent~?v~.~}\\
\lf{?v~st:hasNextVesselEvent~?v1~.~}\\
\lf{?v1~st:hasLocation st:port~.~}\\
\lf{?v1~st:hasTimestamp~?vd~.~}\\
%\lf{BIND(~str(?cd)~as~?cdstr~)~.}\\
\lf{BIND(~fn:substring(?cd,5,10)~AS~?endCI~)~.}\\
%\lf{BIND(~str(?vd)~as~?vdstr~)}\\
\lf{BIND(~fn:substring(?vd,5,10)~AS~?vesStop )~.}\\
\lf{FILTER~(xsd:date(?vesStop)~>~xsd:date(?endCI))~.}\\
\lf{\}}
\end{scriptsize}
 \finepatt
\end{dlquery}

\subsubsection*{Querying \cycle}
%Here we give the corresponding formalisation of pattern \cycle~with OWL-API and  Clark \& Parsia SPARQL-DL syntaxes:
\begin{dlquery}{\bf \cycle in OWL-API}\\
\begin{scriptsize}
\lf{Maritime\_Container\_Itinerary}~\lf{and}~\lf{hasCSourcePort}~\lf{value}~\lf{P1}~\lf{and}\\
~~~~~~~~~~~~\lf{hasCDestinationPort}\\
\lf{value}~\lf{P2}~\lf{and}~\lf{hasContainerEvent}~\lf{some}\\
\lf{(Transhipment\_Event}~\lf{and}~\lf{hasLoadingVesselEvent}~\lf{some}\\
\lf{(hasNextVesselEvent}~\lf{some}~\lf{(Event}~\lf{and}~\lf{hasVPort}~\lf{value}~\lf{P1}~\lf{and}\\
\lf{hasNextVesselEvent}~\lf{some} \lf{(Event}~\lf{and}~\lf{hasVPort}~\lf{value}~\lf{P2))))}
\end{scriptsize}
 \finepatt
\end{dlquery}

\begin{dlquery}{\bf \cycle in SPARQL-DL}\\
\begin{scriptsize}
\lf{SELECT~DISTINCT~?c~?cd~?vd~WHERE~\{~}\\
\lf{?c~a~st:Container\_itinerary~.~}\\
\lf{?c~st:hasEndTime~?cd~.~}\\
\lf{?c~st:hasCISourcePort~st:port1~.~}\\
\lf{?c~st:hasCIDestinationPort~st:port2~.~}\\
\lf{?c~st:hasContainerEvent~?t~.~}\\
\lf{?t~rdf:type~?eventClass~.~}\\
\lf{?eventClass~rdfs:subClassOf~st:Transshipment\_Event~.~}\\
\lf{?t~st:hasLoadingVesselEvent~?v~.~}\\
\lf{?v~st:hasNextVesselEvent~?v1~.~}\\
\lf{?v1~st:hasLocation st:port1~.~}\\
\lf{?v1~st:hasNextVesselEvent~?v2~.~}\\
\lf{?v2~st:hasLocation st:port2~.~}\\
\lf{?v2~st:hasTimestamp~?vd~.~}\\
\lf{\}}
\end{scriptsize}
 \finepatt
\end{dlquery}

%Here it is the alternative formalisation of pattern \cycle~with Clark \& Parsia SPARQL-DL syntax:
\begin{dlquery}{\bf \cycle in SPARQL-DL (alternative formalization)}\\
\begin{scriptsize}
\lf{SELECT~DISTINCT~?c~?endCI~?vesStop~WHERE~\{~}\\
\lf{?c~a~st:Container\_itinerary~.~}\\
\lf{?c~st:hasEndTime~?cd~.~}\\
\lf{?c~st:hasCISourcePort~st:port~.~}\\
\lf{?c~st:hasContainerEvent~?t~.~}\\
\lf{?t~rdf:type~?eventClass~.~}\\
\lf{?eventClass~rdfs:subClassOf~st:Transshipment\_Event~.~}\\
\lf{?t~st:hasLoadingVesselEvent~?v~.~}\\
\lf{?v~st:hasNextVesselEvent~?v1~.~}\\
\lf{?v1~st:hasLocation st:port~.~}\\
\lf{?v1~st:hasTimestamp~?vd~.~}\\
\lf{?v1~st:hasNextVesselEvent~?v2~.~}\\
\lf{?t2~st:hasDischargingVesselEvent~?v2~.~}\\
\lf{?t2~rdf:type~?eventClass2~.~}\\
\lf{?eventClass2~rdfs:subClassOf~st:Transshipment\_Event~.~}\\
\lf{?c~st:hasContainerEvent~?t2~.~}\\
\lf{?t2~st:hasTimestamp~?disDate~.~}\\
\lf{BIND(~fn:substring(?disDate,5,10)~AS~?endvTimeDis)~.~}\\
\lf{BIND(~fn:substring(?cd,5,10) AS~?endCI~)~.~}\\
\lf{BIND( fn:substring(?vdstr,5,10)~AS~?vesStop))~.~}\\
\lf{FILTER (xsd:date(?endCI)~>~xsd:date(?vesStop))~.~}\\
\lf{FILTER (xsd:date(?endvTimeDis)~>~xsd:date(?vesStop))~.~}\\
\lf{\}}
\end{scriptsize}
 \finepatt
\end{dlquery}

% LOOP  loopNewDiscIntermediate performances
%
% 5000:   5m49.588s
% 10000: 7m19.239s
% 15000:  9m8.555s
% 20000:	40m46.455s
%
%
\begin{dlquery}{\bf \cycle in SPARQL-DL (intermediate ports)}\\
\begin{scriptsize}
\lf{SELECT~DISTINCT~?c~?endCI~?vesStop~WHERE~\{~}\\
\lf{?c~a~st:Container\_itinerary~.~}\\
\lf{?c~st:hasEndTime~?cd~.~}\\
%first the container goes to the intermediate location
\lf{?c~st:hasContainerEvent~?interMediate~.~}\\
\lf{?interMediate~st:hasLocation~st:port~.~}\\
\lf{?interMediate~st:hasTimestamp~?interMediateTimeStamp~.~}\\
\lf{?c~st:hasContainerEvent~?t~.~}\\
\lf{?t~rdf:type~?eventClass~.~}\\
\lf{?eventClass~rdfs:subClassOf~st:Transshipment\_Event~.~}\\
%then it is loaded to another vessel that goes back to the intermediate location
\lf{?t~st:hasLoadingVesselEvent~?v~.~}\\
\lf{?v~st:hasNextVesselEvent~?v1~.~}\\
\lf{?v1~st:hasLocation~st:port~.~}\\
\lf{?v1~st:hasTimestamp~?vd~.~}\\
%after that, it is discharged
\lf{?v1~st:hasNextVesselEvent~?v2~.~}\\
\lf{?t2~st:hasDischargingVesselEvent~?v2~.~}\\
\lf{?c~st:hasContainerEvent~?t2~.~}\\
\lf{?t2~rdf:type~?eventClass2~.~}\\
\lf{?eventClass2~rdfs:subClassOf~st:Transshipment\_Event~.~}\\
\lf{?t2~st:hasTimestamp~?disDate~.~}\\
\lf{BIND(~fn:substring(?disDate,5,10)~AS~?endvTimeDis~)~}\\
\lf{BIND(~fn:substring(?interMediateTimeStamp,5,10)~AS~?interMediateTime~).~}\\
\lf{BIND(~fn:substring(?cd,5,10)~AS~?endCI~)~}\\
\lf{BIND(~fn:substring(?vd,5,10)~AS~?vesStop~)~}\\
\lf{FILTER~(xsd:date(?vesStop)~>~xsd:date(?interMediateTime))~.}\\
\lf{FILTER~(xsd:date(?endCI)~>~xsd:date(?vesStop))~.}\\
\lf{FILTER~(xsd:date(?endvTimeDis)~>~xsd:date(?vesStop))~.}\\
\lf{\}}
\end{scriptsize}
 \finepatt
\end{dlquery}

\end{document}